\documentclass[sigconf]{acmart}

\copyrightyear{2024}
\acmYear{2024}
\setcopyright{rightsretained}
\acmConference[SETN 2024]{13th Conference on Artificial Intelligence}{September 11--13, 2024}{Piraeus, Greece}
\acmBooktitle{13th Conference on Artificial Intelligence (SETN 2024), September 11--13, 2024, Piraeus, Greece}\acmDOI{10.1145/3688671.3688747}
\acmISBN{979-8-4007-0982-1/24/09}

\usepackage{natbib}
\usepackage{colortbl}
\usepackage{subcaption}

\newcommand{\term}[1]{\emph{#1\/}}
\newcommand{\quotes}[1]{`#1'}

\begin{document}

\title{Comparing Prior and Learned Time Representations in
  Transformer Models of Timeseries}

\author{Natalia Koliou}
\authornote{Both authors contributed equally to this research.}
\email{{nataliakoliou,tatianabou,konstant}@iit.demokritos.gr}
\orcid{0009-0004-3920-9992}

\author{Tatiana Boura}\authornotemark[1]
\orcid{0009-0008-0656-4372}

\author{Stasinos Konstantopoulos}
\orcid{0000-0002-2586-1726}

\affiliation{%
\institution{Institute of Informatics and Telecommunications, \\
  NCSR \quotes{Demokritos}}
\city{Ag. Paraskevi}
\country{Greece}
}

\author{George Meramveliotakis}
\orcid{0000-0002-0110-347X}
\author{George Kosmadakis}
\orcid{0000-0002-3671-8693}
\email{{gmera,gkosmad}@ipta.demokritos.gr}

\affiliation{%
\institution{Institute of Nuclear \& Radiological Sciences and Technology, Energy \& Safety, \\
  NCSR~\quotes{Demokritos}}
\city{Ag.~Paraskevi}\country{Greece}
}

\begin{abstract}

What sets timeseries analysis apart from other machine learning
exercises is that time representation becomes a primary aspect of the
experiment setup, as it must adequately represent the temporal
relations that are relevant for the application at hand. In the work
described here we study wo different variations of the Transformer
architecture: one where we use the fixed time representation proposed
in the literature and one where the time representation is learned
from the data. Our experiments use data from predicting the energy
output of solar panels, a task that exhibits known periodicities
(daily and seasonal) that is straight-forward to encode in the fixed
time representation. Our results indicate that even in an experiment
where the phenomenon is well-understood, it is difficult to encode
prior knowledge due to side-effects that are difficult to mitigate.
We conclude that research work is needed to work the human into the
learning loop in ways that improve the robustness and trust-worthiness
of the network.

\end{abstract}

\ccsdesc[500]{Computing methodologies~Neural networks}
\ccsdesc[300]{Mathematics of computing~Time series analysis}
\ccsdesc[500]{Computing methodologies~Artificial intelligence}
\ccsdesc[300]{Computing methodologies~Supervised learning}

\maketitle

\section{Introduction}
\label{sec:intro}

What sets apart timeseries analysis from other machine learning
exercises is taking into account the sequence as well as, in most
cases, the temporal distance between observations. This makes the
representation of time a primary aspect of the experiment setup,
as it must be adequate for representing the temporal relations that
are relevant for the application at hand.

To elaborate on the various considerations that need to be addressed,
first consider that one cannot assume fully observed, uniformly
sampled inputs as there might be gaps in the data, varying sampling
rates, and (for multivariate timeseries) misalignment between the
time steps of the different variables. This dictates a representation
that allows time differences to be computed, so that (for example)
September~2023 is \quotes{closer} to January~2024 than it is to
September~2022. Simple timestamps allow this but do not capture
periodicity: Consider, for instance, an application with seasonal
periodicity where September~2023 is \quotes{closer} to
September~2022 than to January~2024.

There is a rich relevant literature in both signal processing and in
non-parametric statistics, as well as in adapting AI/ML approaches to
timeseries processing when facing irregularly sampled and/or sparse data. 
In particular, deep learning approaches that utilize recurrent networks based on Gated Recurrent Units (GRUs)~\cite{che_etal}, Long Short-Term Memory networks (LSTMs)~\cite{pham_etal, neil_etal}, and ODE-RNNs~\cite{mona_etal} have shown promising results.
In the work described we
focus on deep learning methods as well, but specifically on adapting \term{Transformer}
models to timeseries analysis. We will first present how the relevant
literature handles the representation of time when applying
Transformer models to timeseries (Section~\ref{sec:bg}) and then
proceed to propose an alternative representation that is expected
to out-perform the original representation for our specific
application on predicting the energy output of solar panels
(Section~\ref{sec:time}). We close with giving and discussing
comparative experimental results (Section~\ref{sec:exp}) and
conclusions and future work (Section~\ref{sec:conc}).

\section{Background}
\label{sec:bg}

Unlike recurrent and differential equation-based
architectures which process inputs sequentially,
Transformers \citep{og_attention} expect the complete time-series as
input and use the \term{attention} mechanism to look for relationships
between all inputs simultaneously. This has the side-effect that
the temporal order is no longer implied by the order in which the
inputs are presented to the network, so that input vectors must be
augmented with features that represent time. But this also creates
the opportunity to use \term{time embeddings} that represent temporal
information in a way that encodes prior knowledge about the data.

The most characteristic example is periodicity. When the data is
known or suspected to exhibit periodicity,
\term{absolute positional encoding} \citep{wen-etal:2023}
encodes time as two features: the sine and the cosine of the raw
timestamp. This representation allows the Transformer to refer
to both absolute times (since the combination of sine and cosine
maps to a single position on the trigonometric circle) and to a
representation where all (for instance) peak positions have the same
feature value. Referring back to our previous example, remember how we
want September~2022 and September~2023 to be on a similar or identical
time from the periodic perspective, but not from the linear-time
perspective. The Transformer will then have the flexibility to weigh
these features in accordance with the phenomenon being modeled.

The meaningful and successful usage of pre-computed time features requires expert knowledge of the application domain and the periodicities that make sense for the phenomenon being modeled.
In other words, it adds upon the inductive bias of the algorithm.

As expected, a line of research emerged that aims to automatically
acquire the time representation from the data.
The \term{multi-time attention network (mTAN)} \citep{mTAN} is a
Transformer model enhanced with a time-attention module. This module
learns temporal patterns by leveraging a trainable embedding function
$\phi(t)$ that maps time $t$ (scaled to 0..1) to a vector of length $d$.

This vector has one linear element that captures the linear
progression of time and $d-1$ sinusoidal elements that capture 
different periodicities as frequency/phase pairs. Formally, the
embedding function is:
\begin{equation}
\phi_i(t) =
\begin{cases}
    \omega_{0} t + \alpha_{0}, & \text{if } i = 0, \\
    \sin(\omega_{i} t + \alpha_{i}), & \text{if } 0<i<d
\end{cases}
\end{equation}
\\
where $\omega_{i}$ and $\alpha_{i}$ are the learnable parameters.

In concrete neural network terms, this is implemented as one linear
layer for $\omega_0,\alpha_0$
and one linear layer with a sine activation function for all other
$\omega_i,\alpha_i$. More precisely, it is implemented with multiple
such structures, one for each attention head. It should be noted
at this point that for the matrix multiplications to work out, this
modeling has the side-effect that $d$ must be the same as the
dimensionality of the output vector.

This modeling gives the Transformer the flexibility to search for 
the period/phase pairs that best capture the periodicities implied
by the data. Only minimal bias is introduced, namely that time is
structured as a sine/linear pair.

\section{Representing Time} \label{sec:time}

\begin{figure}[tb]
    \centering
    \includegraphics[width=\linewidth]{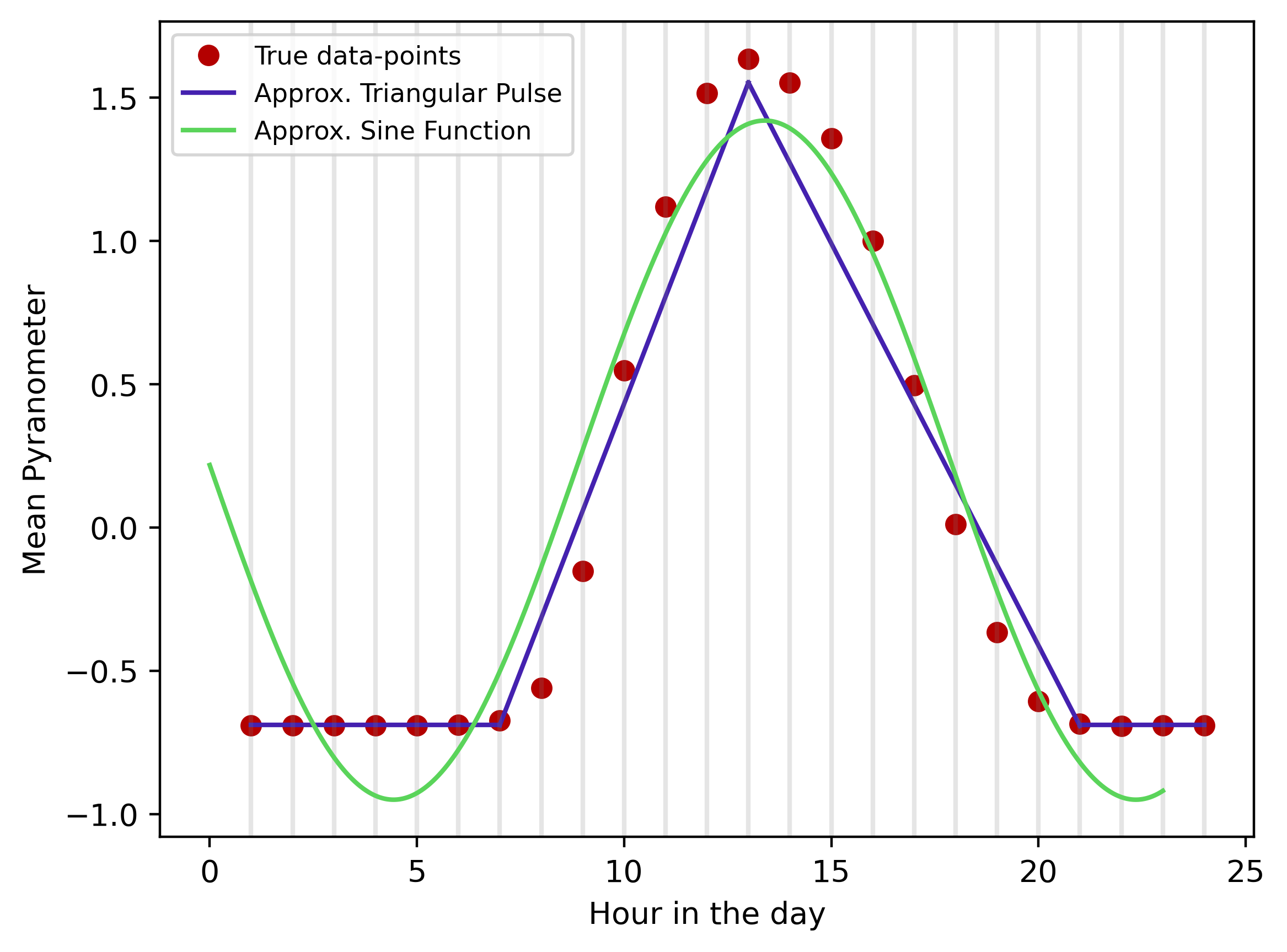}
    \caption{Hourly mean pyranometer values (red) and their function approximations, triangular pulse (blue) and sinusoidal function (green).}
    \label{fig:mean_pyranometer_approx}
\end{figure}

In order to experiment with time representations and the effect they
have on the quality of the learned model, we have assumed an
application where we need to predict the energy output of solar
thermal collectors based on external conditions. More detailed
information about the application and the data is provided in
Section~\ref{sec:exp}, and for the purposes of the current discussion
it suffices to mention that the input variables are solar radiation
and external temperature, which exhibit the obvious daily and annual
periodicities.

We have noted however, that solar radiation behaves in a way that is
not captured by sinusoidal functions. While these functions represent
well the similarity between the same time on a different day, as well
as the similarity between early morning and late afternoon, they fail
to represent the fact that the whole of the night-time is the same and
(as far as solar radiation is concerned) it makes no difference if the
time is 10pm or 24am. We have, therefore, thought that a triangular
spike will be a better representation than a sine. This is further
corroborated by the data shown in
Figure~\ref{fig:mean_pyranometer_approx}, showing pyranometer (solar
radiation) values aggregated to 1h intervals and averaged through the year,
and the best approximation that can be achieved by the sine function and
by the triangular spike.

Based on the argument above, we have defined the four alternative time
embeddings presented below.

\subsection{Triangular Pulse \& Linear}

The first embedding is a season-modulated triangular pulse/linear
pair. The base and peak of the triangular pulse is not the same for
each day, but is calculated so that the pulse will start at sunrise,
peak at noon, and end at sunset. This represents that solar radiation
at noon is distinct from all other times during day, one hour before
noon is similar to one hour after noon, and so on until
sunrise/sunset. Outside the base of the pulse, all times are represented
by the same value of 0.01 to denote that the distinction between them
is not important.\footnote{This value is set to 0.01 instead of 0 to
  satisfy a technical requirement of the implementation.}

As for the linear function, it provides a straightforward method for ensuring the 
uniqueness of timestamps within the representation. By mapping each timestamp to 
a unique value, we establish clear distinctions between different time points. 
Unlike the periodicity of the triangular pulse, the linear function spans the 
entire range of timestamps, from the earliest to the latest, using both date and 
time components. This continuous representation does not reset daily but maps the 
entire period linearly, ensuring a unique value for each timestamp based on its 
position in the overall time span.

The second embedding is similar to the one above, but the pulse is fixed
to start at 7am, peak at 1pm, and end at 9pm regardless of the date.
This simpler embedding is included in order to be able to see if the
more complex approach above has added value or there is no
significant loss in accuracy when using fixed parameters.

\subsection{Sinusoidal}

The third embedding follows the absolute positional encoding
literature and consists of two sine/cosine pairs, one pair computed
from the month of the year and one pair computed from the hour of the
day. Sine and cosine capture cyclical patterns effectively, and by
combining the hour and the month embedding we guarantee that the
Transformer has the means to model daily and annual periodicities and
also to refer to absolute timepoints if the data prove this useful.

To find the appropriate phase shift and period, we argue as follows:
To represent the daily cycle we need our sine wave to peak at noon. We
aim for noon to be distinct, while the remaining timestamps should
exhibit symmetrical correlation. To achieve this, observe how a
period of 12 for hours and 24 for months, with a zero shift, will work
as expected. Naturally, the cosine must be parameterized with identical
phase and period to maintain the property that the sine/cosine pair
uniquely refers to a point in the trigonometric circle.

\subsection{Sine \& Sawtooth}

The fourth and final embedding consists of a sine and a sawtooth
function. Just like in the sinusoidal embedding, we also use the sine wave with
the same parameters (shift, period) to express
correlation among timestamps. However, when it comes to expressing uniqueness, we considered using 
a sawtooth wave instead of a cosine wave, to observe whether any noticeable changes might take place. 
Unlike the linear function used in the triangular pulse embeddings which spans the entire range of timestamps, the sawtooth 
wave resets to 0 output after each day. The 
sawtooth wave parameters (shift, period) are set to (6, 6) for hours and (12, 12) for months to scale the
 output values for both cases within [-1, 1].

\begin{figure}[tbp]
    \centering
    \includegraphics[scale=0.6]{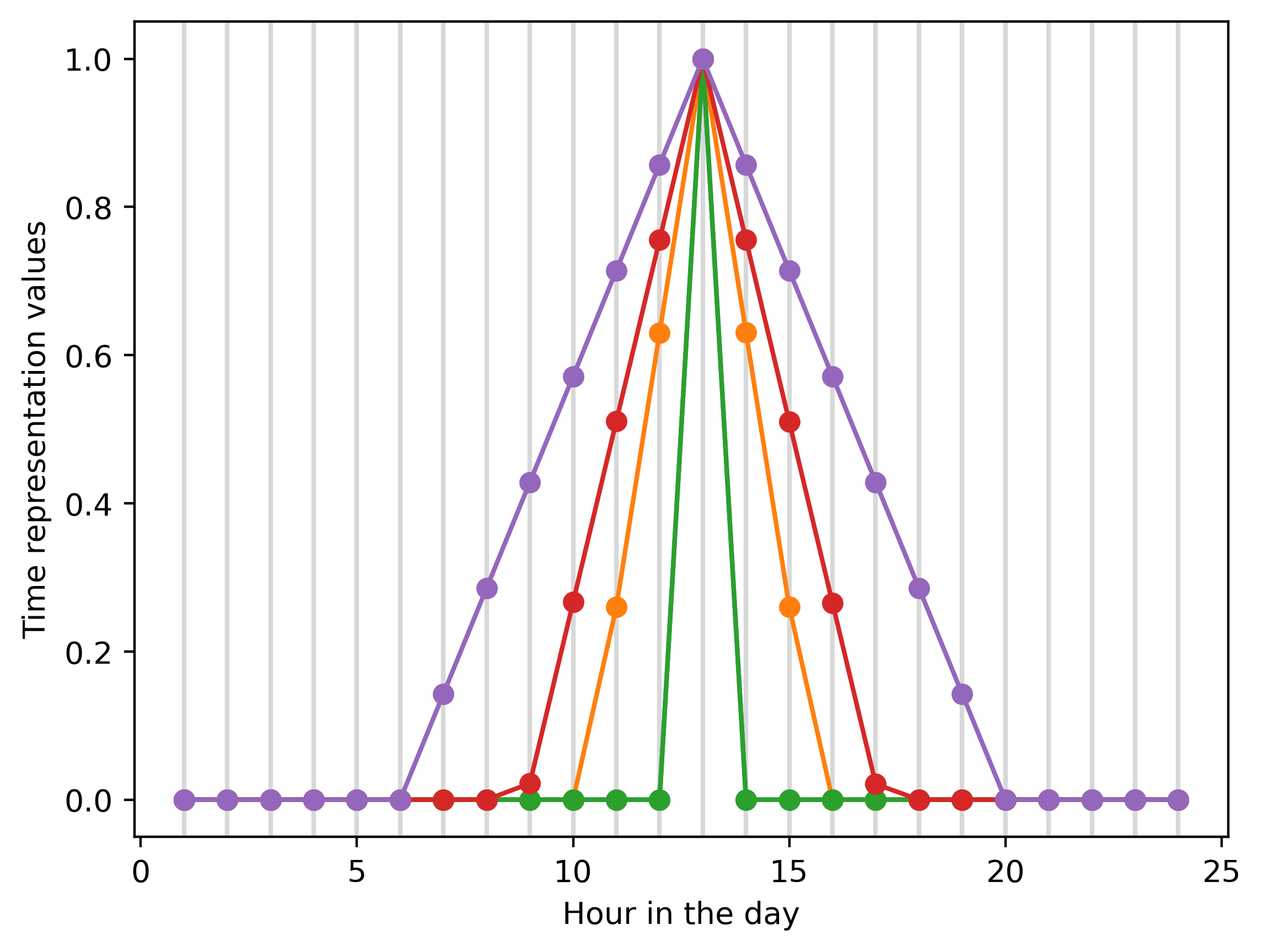}
    \caption{Examples of learned time representation features with a learnable triangular pulse. The five illustrated time features come from the same testing sample and highlight the capability of the model to learn (non-)isosceles pulses with different bases.}
    \label{fig:learned_pulse}
\end{figure}

\subsection{Learned Time Representations}

As mentioned in the Background, the idea behind the \term{mTAN} is to learn periodic features with a sinusoidal representation.
We have argued that, for our application, a triangular pulse for representing the non-linear features may be a better fit. 
Thus, we altered the activation function of the presented model to convert the linear layer to a triangular pulse.

Our first attempt at creating such a function involved using a triangular function with a fixed base. 
In this case, the start and end of the pulse corresponded to the hours when difference of the solar output peaked (7 AM and 9 PM respectively).
This approach was not fruitful, as the model was not given the opportunity to learn the periodic representations.

Naturally, we then focused on learning the base of the pulse. 
Choosing 1 PM as the peak of the pulse, the two most straightforward approaches we implemented were: 
(a) splitting the time vector and then using the hours where the absolute difference of the elements in the learned time vector is the largest as the start and end of the pulse; and 
(b) splitting the time vector and then using the hours where the elements themselves in the learned time vector have the absolute largest values as the start and end of the pulse.
However, both implementations failed to learn any meaningful form of pulse and remained fixed on the initial time parameters.

The approach that successfully achieved the task of learning different triangular pulses was an engineering one. It emerged from the idea that the non-linear function itself should be simple in terms of numerical computation and traceability, since the previous, more complex approaches were not suitable for the task.
The idea is described as follows: we first calculate the absolute difference of each value in the time vector of each representation of size \(d\) and the value of the \(13^{\text{th}}\) element of this representation (corresponding to 1 PM). Then we compute the \(25^{\text{th}}\) percentile of these differences and replace each difference \(\text{dist}_{ij}\) with zero if it is less than the percentile value \(v_i\) of the corresponding representation \(i\), and with \(1 - \frac{\text{dist}_{ij}}{v_i}\) otherwise.
Figure \ref{fig:learned_pulse} presents a few example triangular pulse time representations learned from employing this approach.

\section{Experiments and Results} \label{sec:exp}

\begin{table*}[tb]
\centering
\caption{Evaluation results for the classification task using prior and learned time representations. Each model was trained using six different initialization seeds, and the final metrics are composed from the aggregation (mean and standard deviation) of the results from these trained models.}
\scalebox{0.9}{
	\begin{tabular}{
		lcccc@{\hspace{0.6cm}}cc
	}
	\toprule
	& \multicolumn{4}{c}{\textbf{Prior Time Representations}} & \multicolumn{2}{c}{\textbf{Learned Time Representations}} \\
	\midrule
	Metric & tr. pulse \& linear $\dagger$ & fixed tr. pulse \& linear $\dagger$ & sine \& cosine $\ddagger$ & sine \& sawtooth $\ddagger$ & sine \& linear $\dagger$ & tr. pulse \& linear $\dagger$ \\
	\midrule
	Precision (Micro) & $0.835 \pm 0.017$ & $0.823 \pm 0.013$ & $\mathbf{0.866} \pm 0.010$ & $0.860 \pm 0.004$ & $\mathbf{0.842} \pm 0.009$ & $0.778 \pm 0.010$ \\
	Recall (Micro) & $0.835 \pm 0.017$ & $0.823 \pm 0.013$ & $\mathbf{0.866} \pm 0.010$ & $0.860 \pm 0.004$ & $\mathbf{0.842} \pm 0.009$ & $0.778 \pm 0.010$ \\
	\rowcolor{blue!10} F-score (Micro) & $0.835 \pm 0.017$ & $0.823 \pm 0.013$ & $\mathbf{0.866} \pm 0.010$ & $0.860 \pm 0.004$ & $\mathbf{0.842} \pm 0.009$ & $0.778 \pm 0.010$ \\
	Precision (Macro) & $0.650 \pm 0.027$ & $0.629 \pm 0.012$ & $\mathbf{0.679} \pm 0.025$ & $0.666 \pm 0.009$ & $\mathbf{0.647} \pm 0.012$ & $0.582 \pm 0.015$ \\
	Recall (Macro) & $0.715 \pm 0.033$ & $0.673 \pm 0.015$ & $\mathbf{0.722} \pm 0.024$ & $0.721 \pm 0.026$ & $\mathbf{0.694} \pm 0.023$ & $0.621 \pm 0.016$ \\
	\rowcolor{blue!10} F-score (Macro) & $0.663 \pm 0.030$ & $0.632 \pm 0.013$ & $\mathbf{0.689} \pm 0.019$ & $0.676 \pm 0.016$ & $\mathbf{0.646} \pm 0.019$ & $0.571 \pm 0.013$ \\
	Precision (Weighted) & $0.884 \pm 0.010$ & $0.876 \pm 0.006$ & $0.894 \pm 0.009$ & $\mathbf{0.895} \pm 0.004$ & $\mathbf{0.892} \pm 0.005$ & $0.867 \pm 0.007$ \\
	Recall (Weighted) & $0.835 \pm 0.017$ & $0.823 \pm 0.013$ & $\mathbf{0.866} \pm 0.010$ & $0.860 \pm 0.004$ & $\mathbf{0.842} \pm 0.009$ & $0.778 \pm 0.010$ \\
	\rowcolor{blue!10} F-score (Weighted) & $0.853 \pm 0.013$ & $0.842 \pm 0.008$ & $\mathbf{0.876} \pm 0.009$ & $0.872 \pm 0.004$ & $\mathbf{0.860} \pm 0.007$ & $0.812 \pm 0.009$ \\
	\bottomrule
	\multicolumn{7}{l}{\small $\dagger$: input being sec/min/hour/day/month/year \quad $\ddagger$: input being month/hour}
\end{tabular}

}
\label{tab:eval_scores}

\end{table*}

\begin{figure}[tb]
    \centering
    \includegraphics[width=\linewidth]{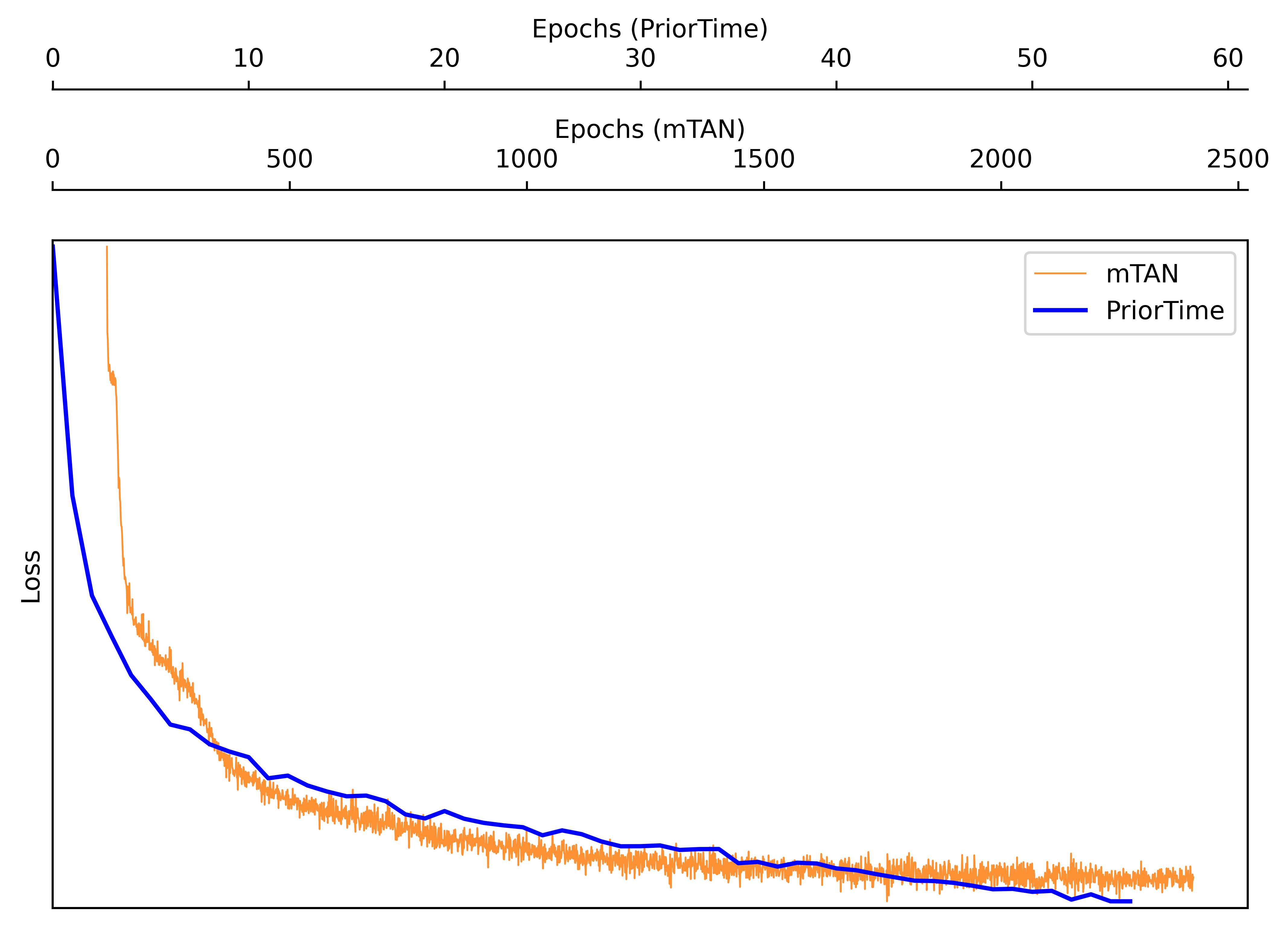}
    \caption{Normalized training loss progression for each model. The models presented are the best-performing models across different time representations (prior and learned). }
    \label{fig:losses}
\end{figure}

\begin{figure*}[ht]
    \centering
    \begin{subfigure}[b]{0.48\textwidth}
        \centering
        \includegraphics[width=\textwidth]{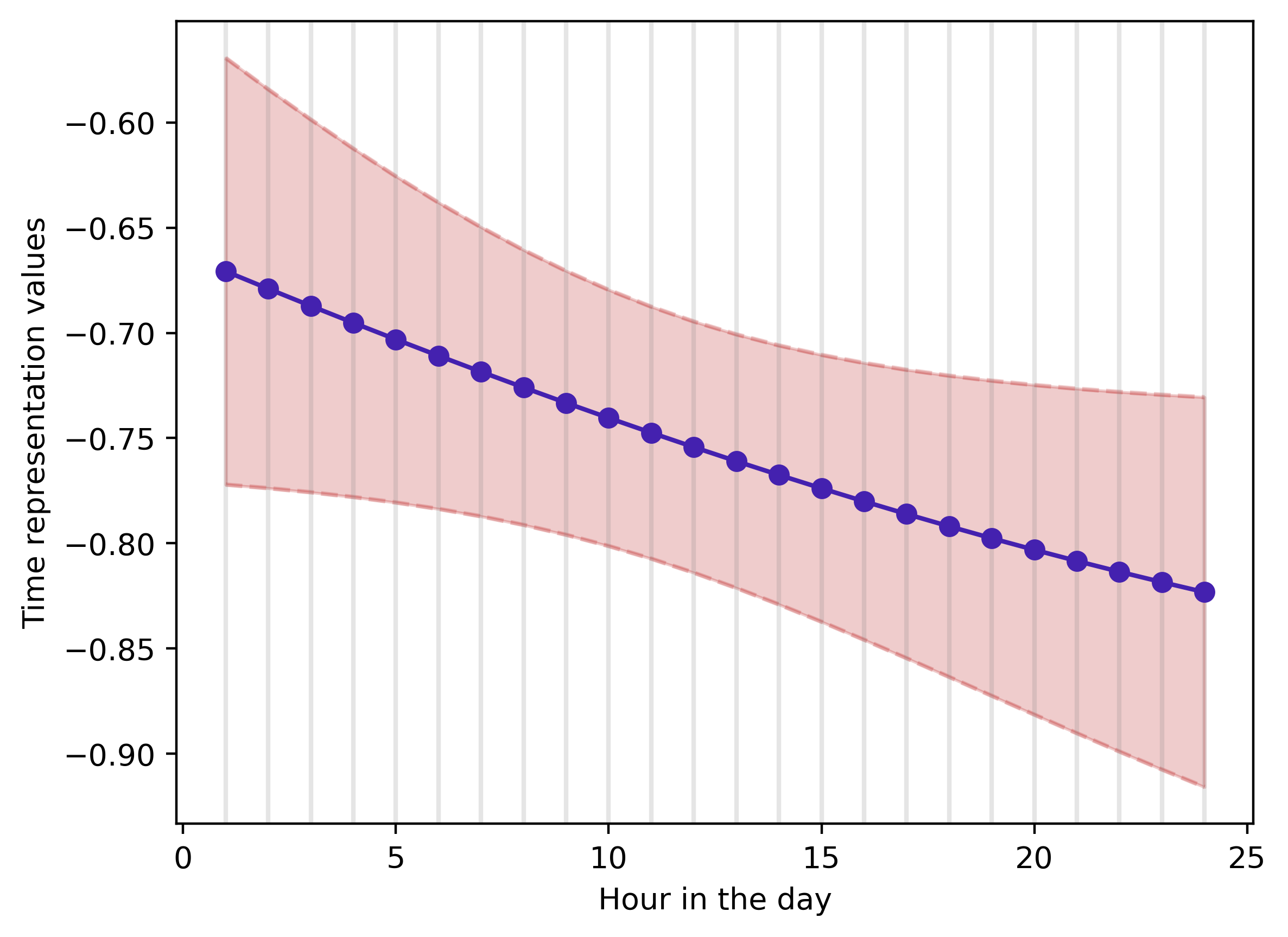}
        \caption{Initial feature representation}
        \label{fig:init_sin}
    \end{subfigure}
    \hfill
    \begin{subfigure}[b]{0.48\textwidth}
        \centering
        \includegraphics[width=\textwidth]{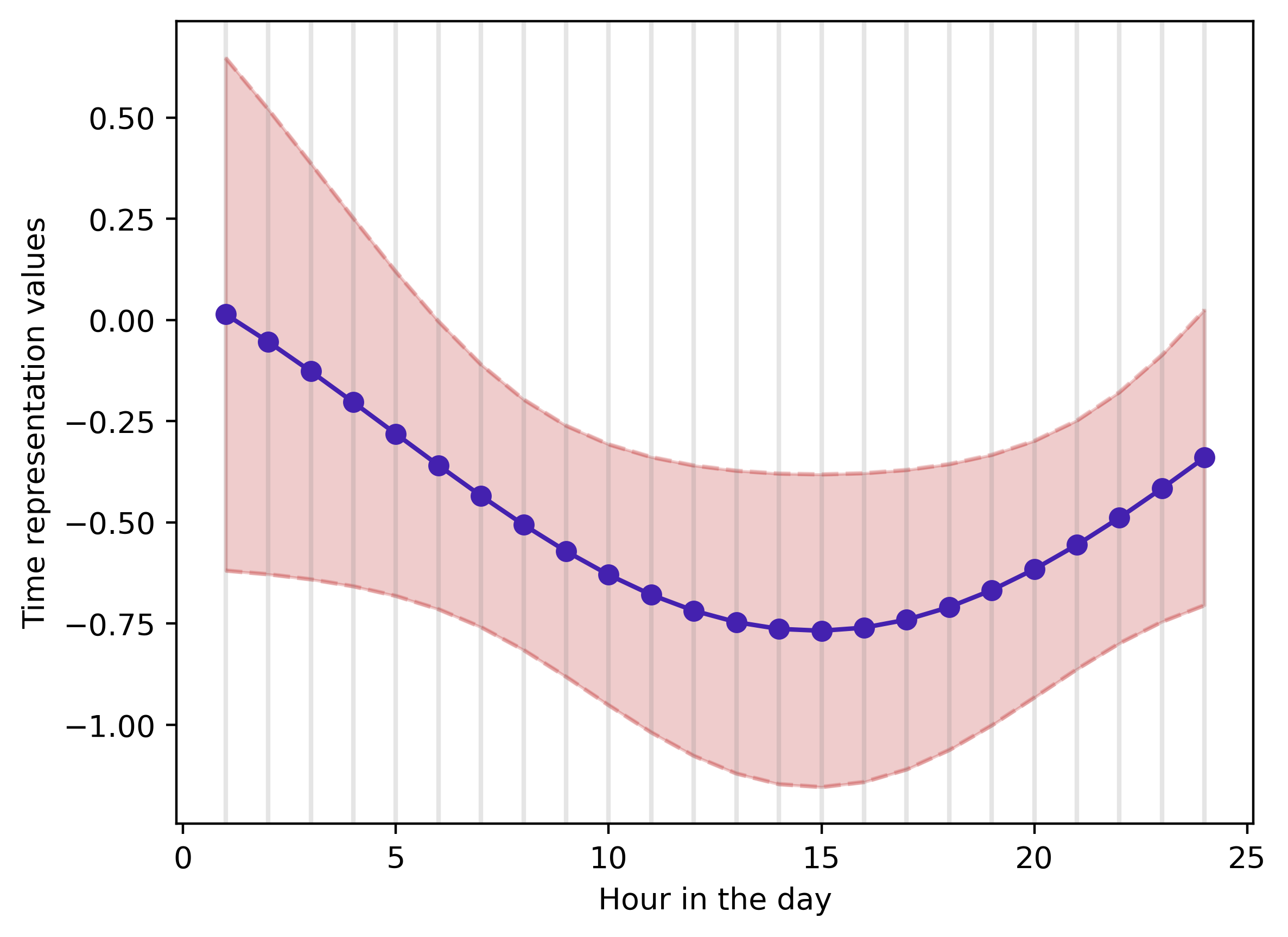}
        \caption{Final feature representation}
        \label{fig:final_sin}
    \end{subfigure}
    \caption{Example of the learning progress of a sine time feature. Both figures illustrate the time representation of the test examples by aggregating them per time-feature using the mean and standard deviation. The left figure shows the feature before the learning process starts, while the right one shows the final learned feature. The learned time representation is meaningful since it distinguishes midday and has similar values for hours with similar accumulated daylight. }
    \label{fig:sine_b4_and_after}
\end{figure*}

\subsection{Experimental setup}

Our data was collected from a pilot building at NCSR
\quotes{Demokritos}. The pilot building features
solar thermal collectors used to experiment
with how to most efficiently control heating systems (solar, heat
pump, etc.) to achieve satisfactory space heating, satisfy hot water
demand, and minimize electric power consumption. The machine learning
task associated with this application is to predict solar power
production from variables reflecting external conditions, namely solar
radiation and external temperature. The output of the model is not
specific power in terms of kWh, but a label that characterizes power
production as being in one of five classes, defined based on expected
demand.

Based on the above, the machine learning task is to transform a
multi-variate timeseries of external conditions aggregated into
one-hour intervals, into a timeseries of power production levels
(the five-class labeling schema mentioned above). The dataset has
a moderate amount of gaps due to the sensors occasionally giving
erroneous, out-of-range readings which are removed. The dataset
exhibits the obvious daily periodicity, and also exhibits a
non-periodic trend. The non-periodic trend is in reality seasonal
periodicity, but since we have used the data from 10 months it
appears to the machine learning task as a non-periodic
trend.\footnote{Available at
  \url{https://zenodo.org/records/12818885}}

To apply the prior time representation we implemented a Transformer
with an encoder, a decoder, and a classifier, in that order. We refer
to this architecture as \term{PriorTime} in the results presented below.
The encoder maps the input into a latent representation, the decoder
reconstructs it preserving the original dimensionality of its
features, and the classifier (a single linear layer) outputs a
probability distribution over the five labels.
To apply the learned time representation, we used the \term{mTAN}
architecture, where the decoder is the same single-layer linear
classifier.
Figure~\ref{fig:losses} give the training loss from training
these models. The complete experimental setup is also
published.\footnote{See guide under \texttt{time\_representations/installation}
  at the \texttt{AINST-2024} branch of repository
  \url{https://github.com/data-eng/navgreen} and also directly accessible at
  \url{https://github.com/data-eng/navgreen/releases/tag/AINST-2024}}

\subsection{Classification results and discussion}

To evaluate the performance of our models on the classification task, we compared 
both the \term{PriorTime} and \term{mTAN} models across different time representations. Our 
evaluation criteria included various performance metrics such as precision, recall,
 and F1-score, as shown in Table \ref{tab:eval_scores}.

The \term{PriorTime} model was trained and tested using four different time representations: 
triangular pulse \& linear, fixed triangular pulse \& linear, sine \& cosine, and sine \& 
sawtooth. Among these representations, the sine \& cosine approach consistently demonstrated 
the strongest performance. The sine \& sawtooth representation approached the results achieved 
by sine \& cosine; however, it only outperformed once, and the
difference was not significant. Concerning the two triangular pulse/linear representations, 
they exhibited the weakest performance overall, with the fixed variation performing slightly 
worse than the season-modulated one. This was not surprising, as we expected the more informed 
time representation to give better results.

The second point is that the sine representations consistently
outperform the triangular ones in both settings (prior and learned),
despite the fact that the triangular pulse is more precise
when it comes to making similar times have similar inputs (solar
radiation and temperature), as already discussed in the beginning
of Section~\ref{sec:time}. We believe that this is due to the fact
that the identical close-to-zero values outside the base of the pulse
hinder back-propagating loss to the linear layer that feeds the
sine/triangular activation function.

When these two points are put together, they imply that
(a) the first linear layer is capable of managing both seasonal
variation and the daily periodicity by properly weighting the inputs
of the activation functions;
(b) (almost) zeroing-out inputs hinders back-propagation; and
(c) having an activation function that dynamically changes to model the 
unique characteristics of each day individually enhances convergence.

\subsection{Discussion of the Learned Time Representations}

\begin{figure*}[tp]
    \centering
    \begin{subfigure}[b]{0.32\textwidth}
        \centering
        \includegraphics[width=\textwidth]{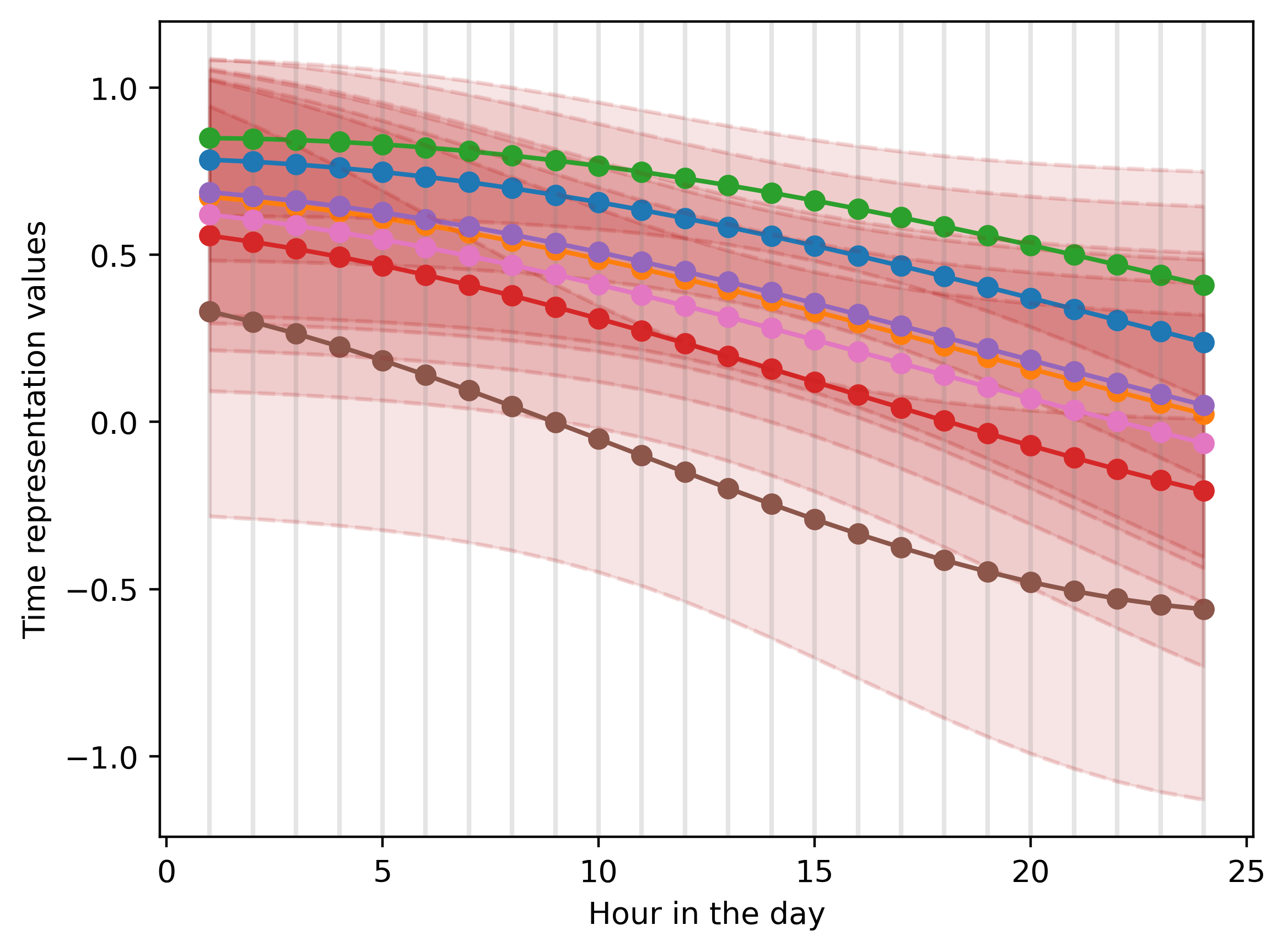}
        \caption{Group 1 (7 time representations)}
        \label{fig:grp1}
    \end{subfigure}
    \hfill
    \begin{subfigure}[b]{0.32\textwidth}
        \centering
        \includegraphics[width=\textwidth]{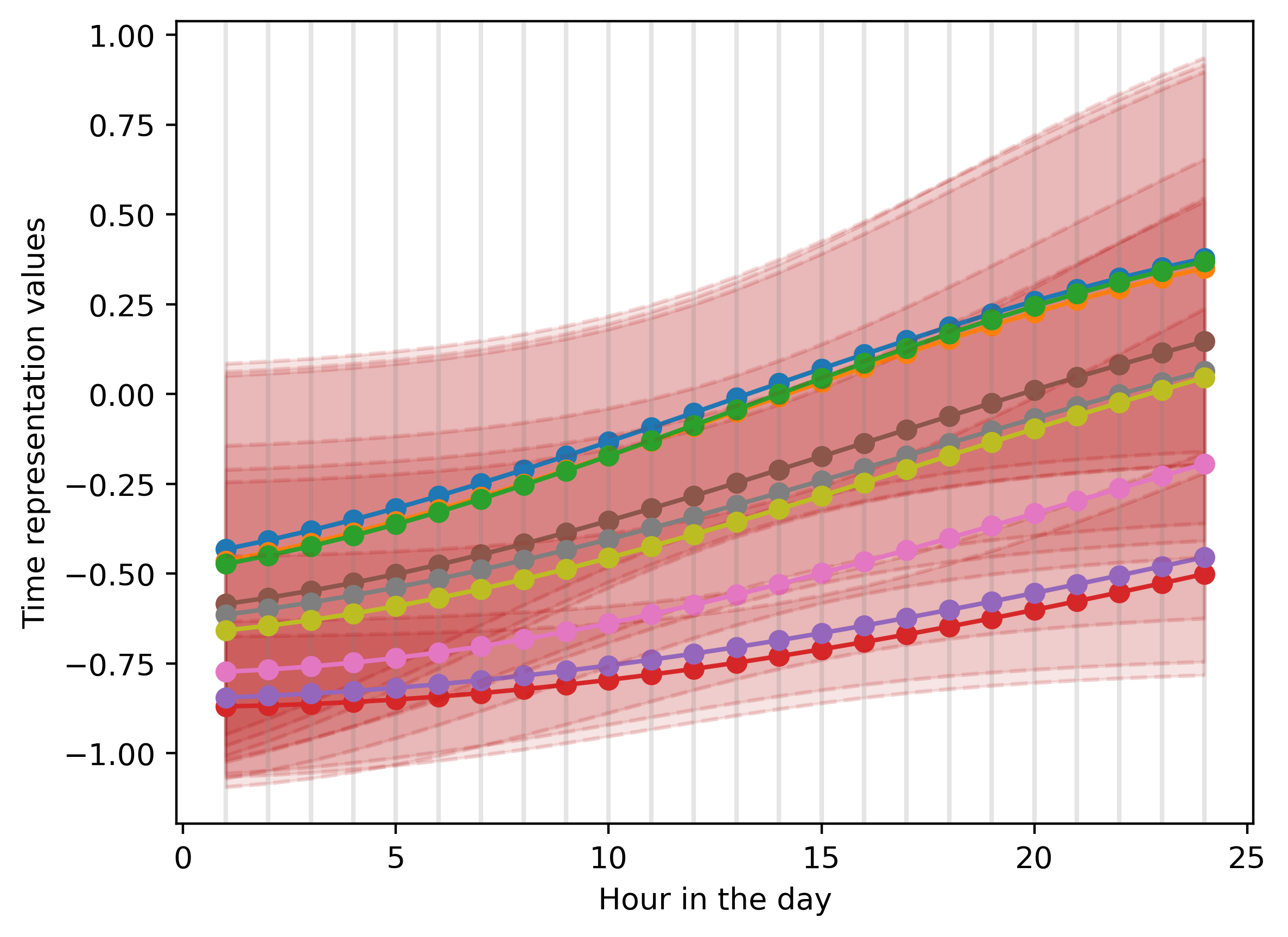}
        \caption{Group 2 (9 time representations)}
        \label{fig:grp2}
    \end{subfigure}
    \hfill
    \begin{subfigure}[b]{0.32\textwidth}
        \centering
        \includegraphics[width=\textwidth]{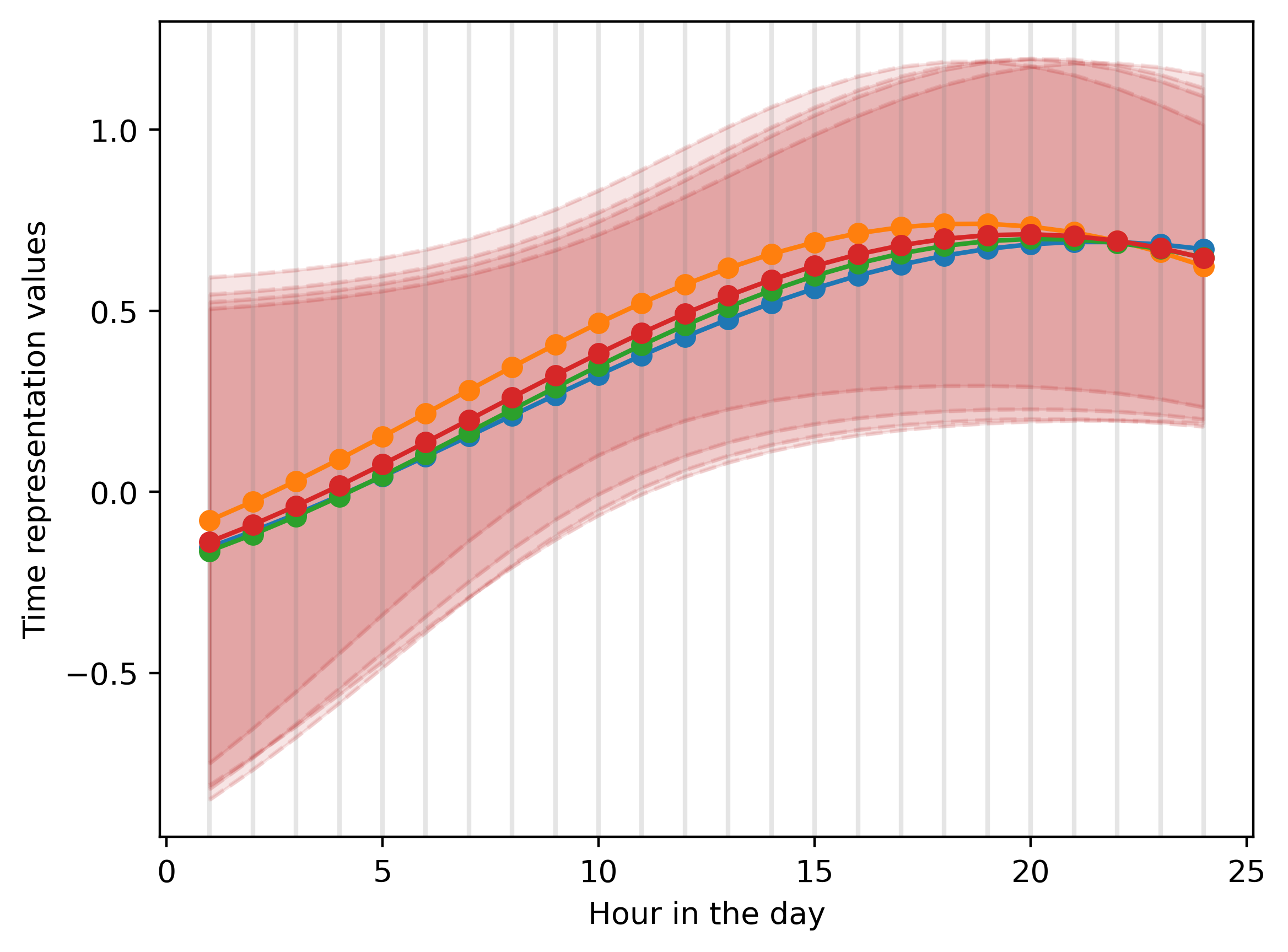}
        \caption{Group 3 (4 time representations)}
        \label{fig:grp3}
    \end{subfigure}
    \vfill
    \begin{subfigure}[b]{0.32\textwidth}
        \centering
        \includegraphics[width=\textwidth]{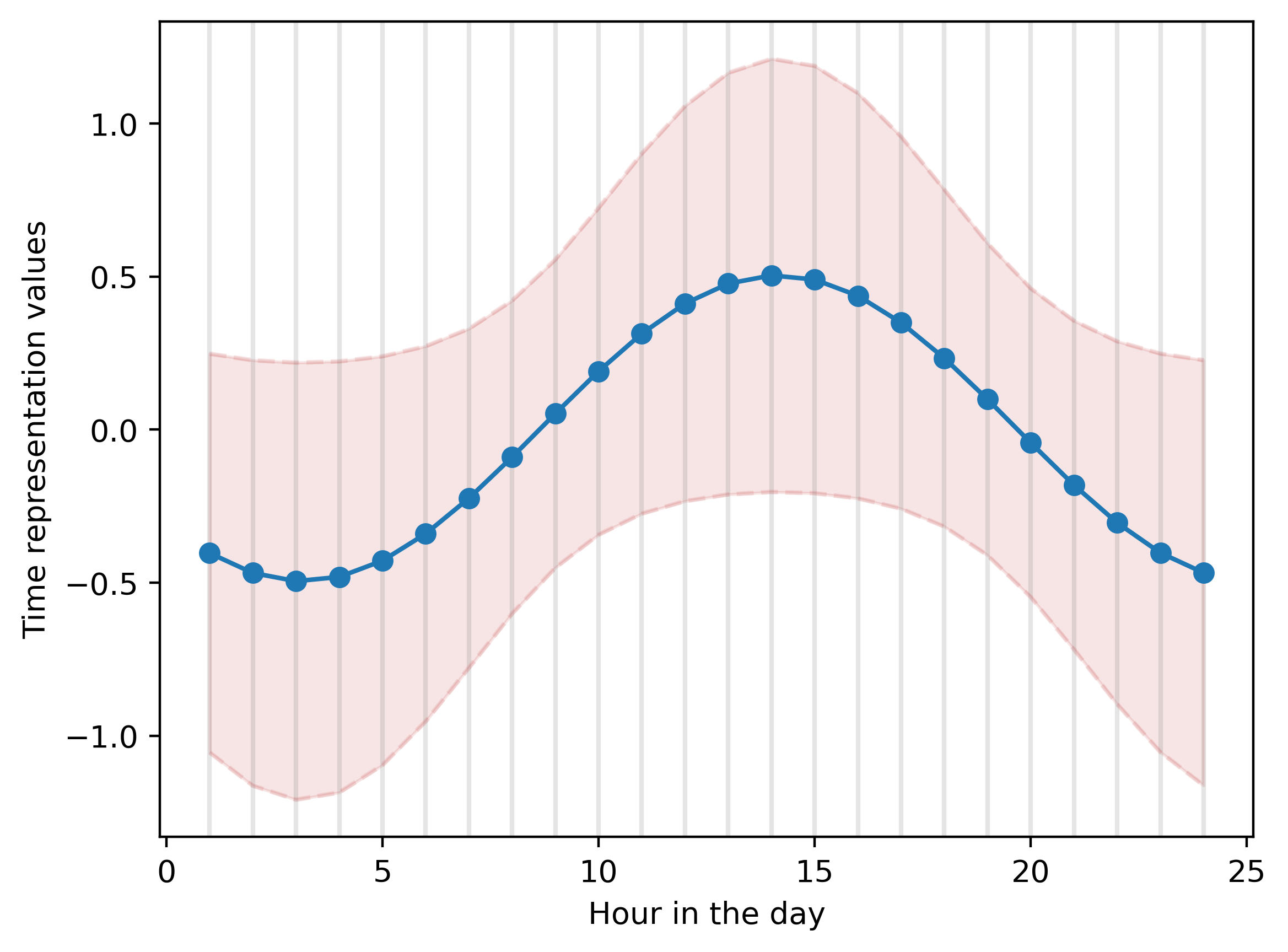}
        \caption{Group 4 (1 time representation)}
        \label{fig:grp4}
    \end{subfigure}
    \hfill
    \begin{subfigure}[b]{0.32\textwidth}
        \centering
        \includegraphics[width=\textwidth]{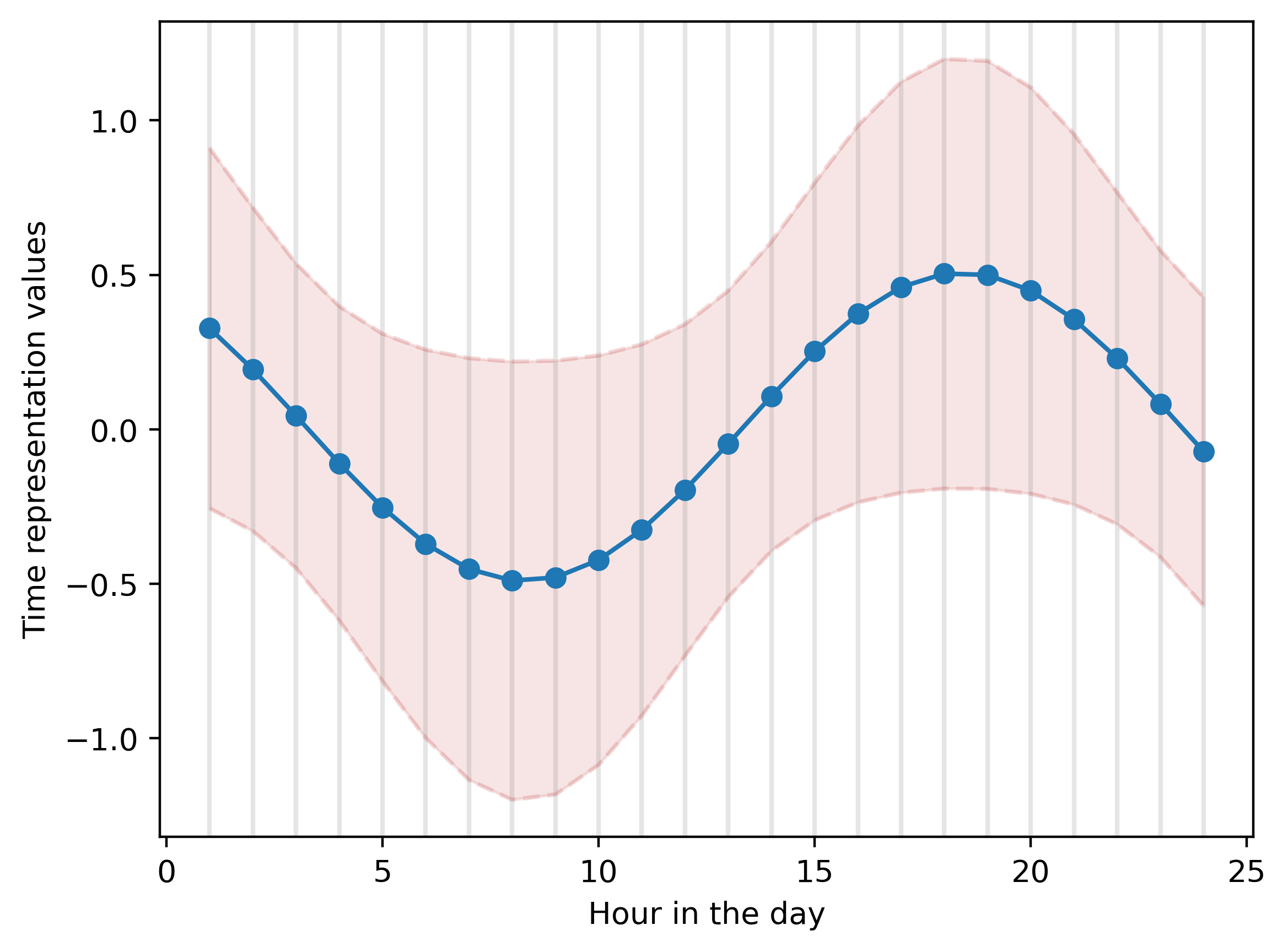}
        \caption{Group 5 (1 time representation)}
        \label{fig:grp5}
    \end{subfigure}
    \hfill
    \begin{subfigure}[b]{0.32\textwidth}
        \centering
        \includegraphics[width=\textwidth]{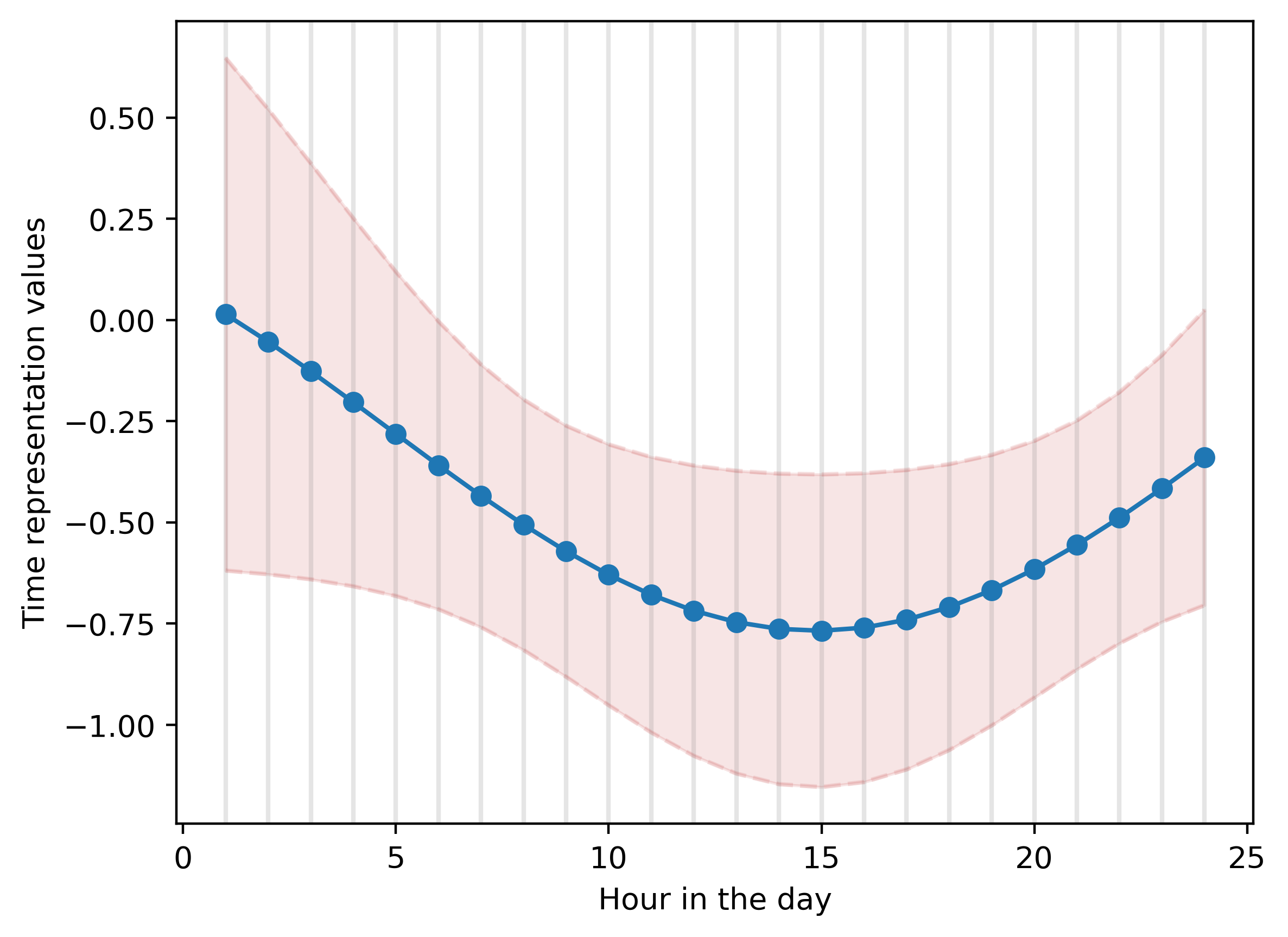}
        \caption{Group 6 (1 time representation)}
        \label{fig:grp6}
    \end{subfigure}
    \caption{Learned sine time representations grouped by their similarity. Each figure is comprised of the aggregated representations computed on the testing data (mean and standard deviation of each time-feature). The top three figures gather the majority of the features, whereas the bottom three include only one feature each. }
    \label{fig:grouped_sines}
\end{figure*}

\begin{figure*}[tp]
    \centering
    \begin{subfigure}[b]{0.32\textwidth}
        \centering
        \includegraphics[width=\textwidth]{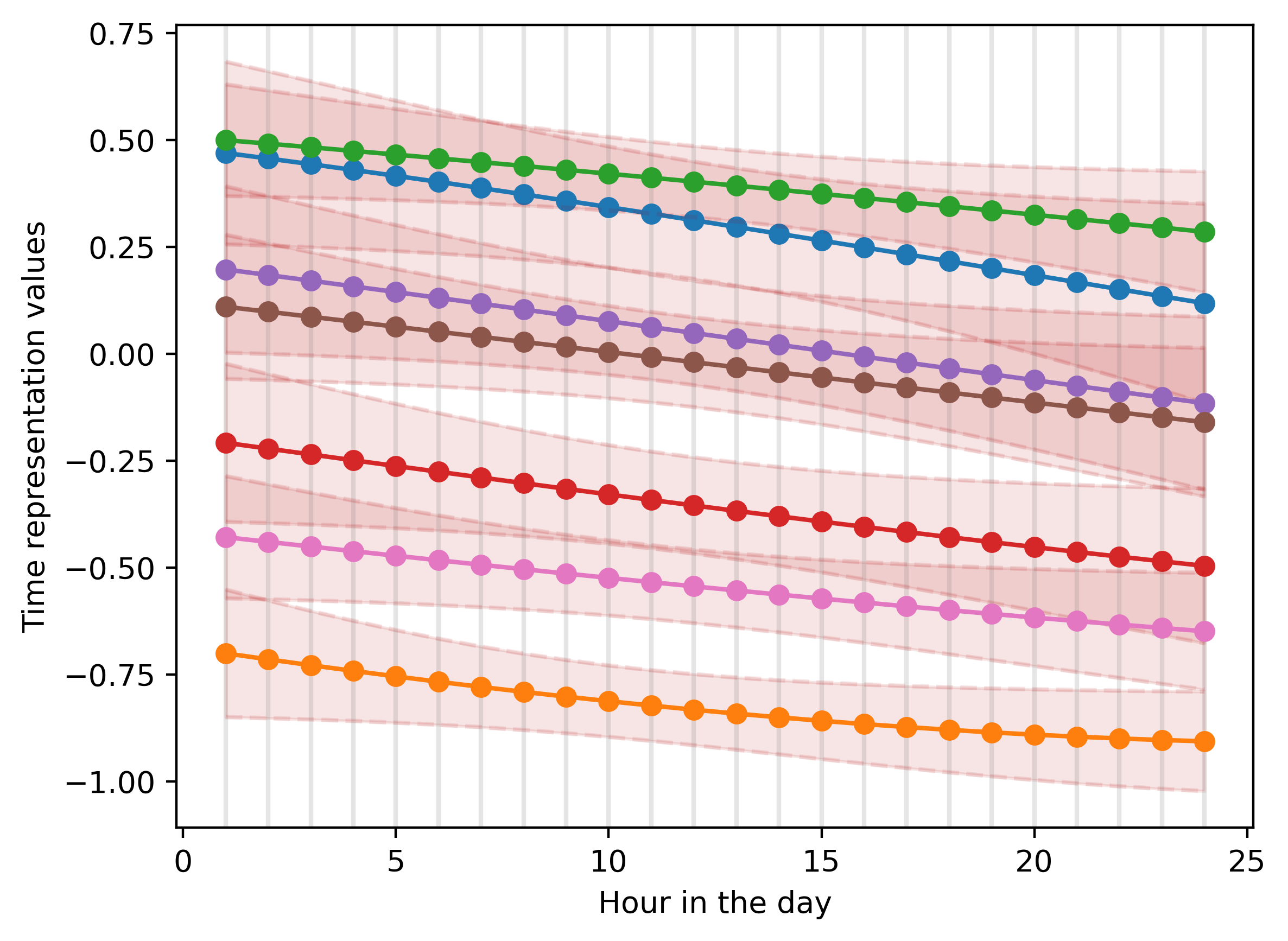}
        \caption{Group 1}
        \label{fig:grp1_init}
    \end{subfigure}
    \hfill
    \begin{subfigure}[b]{0.32\textwidth}
        \centering
        \includegraphics[width=\textwidth]{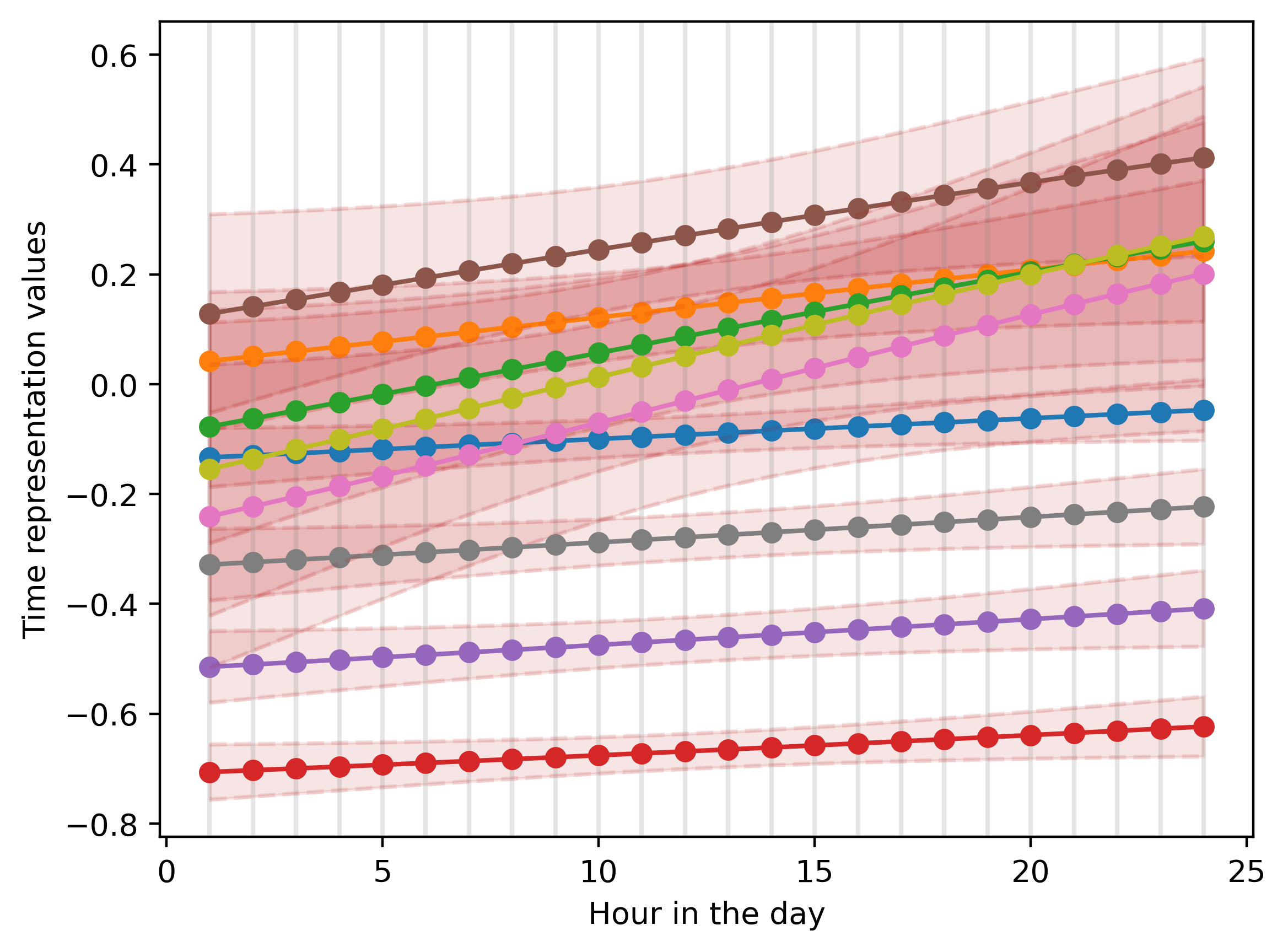}
        \caption{Group 2}
        \label{fig:grp2_init}
    \end{subfigure}
    \hfill
    \begin{subfigure}[b]{0.32\textwidth}
        \centering
        \includegraphics[width=\textwidth]{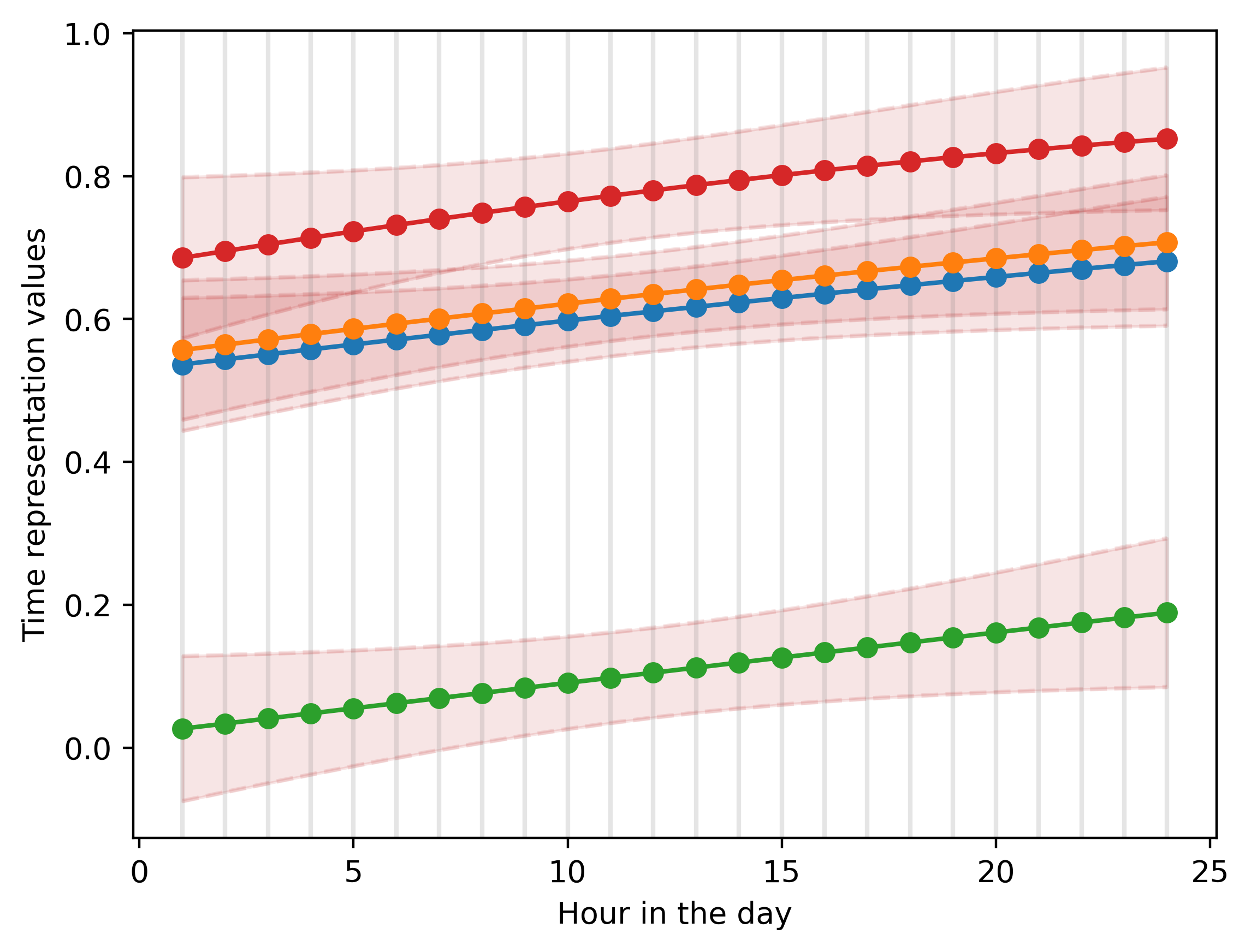}
        \caption{Group 3}
        \label{fig:grp3_init}
    \end{subfigure}
    \vfill
    \begin{subfigure}[b]{0.32\textwidth}
        \centering
        \includegraphics[width=\textwidth]{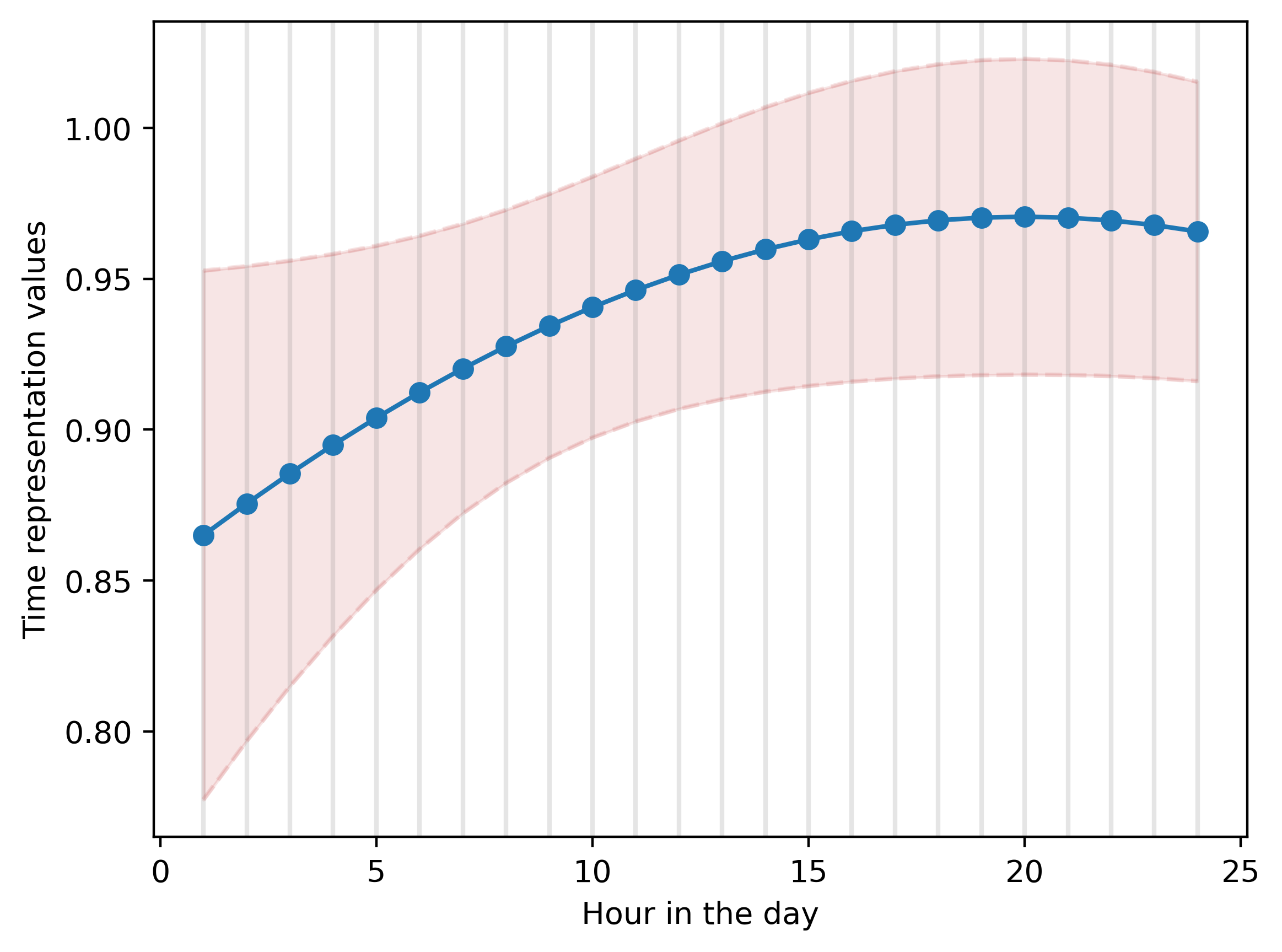}
        \caption{Group 4}
        \label{fig:grp4_init}
    \end{subfigure}
    \hfill
    \begin{subfigure}[b]{0.32\textwidth}
        \centering
        \includegraphics[width=\textwidth]{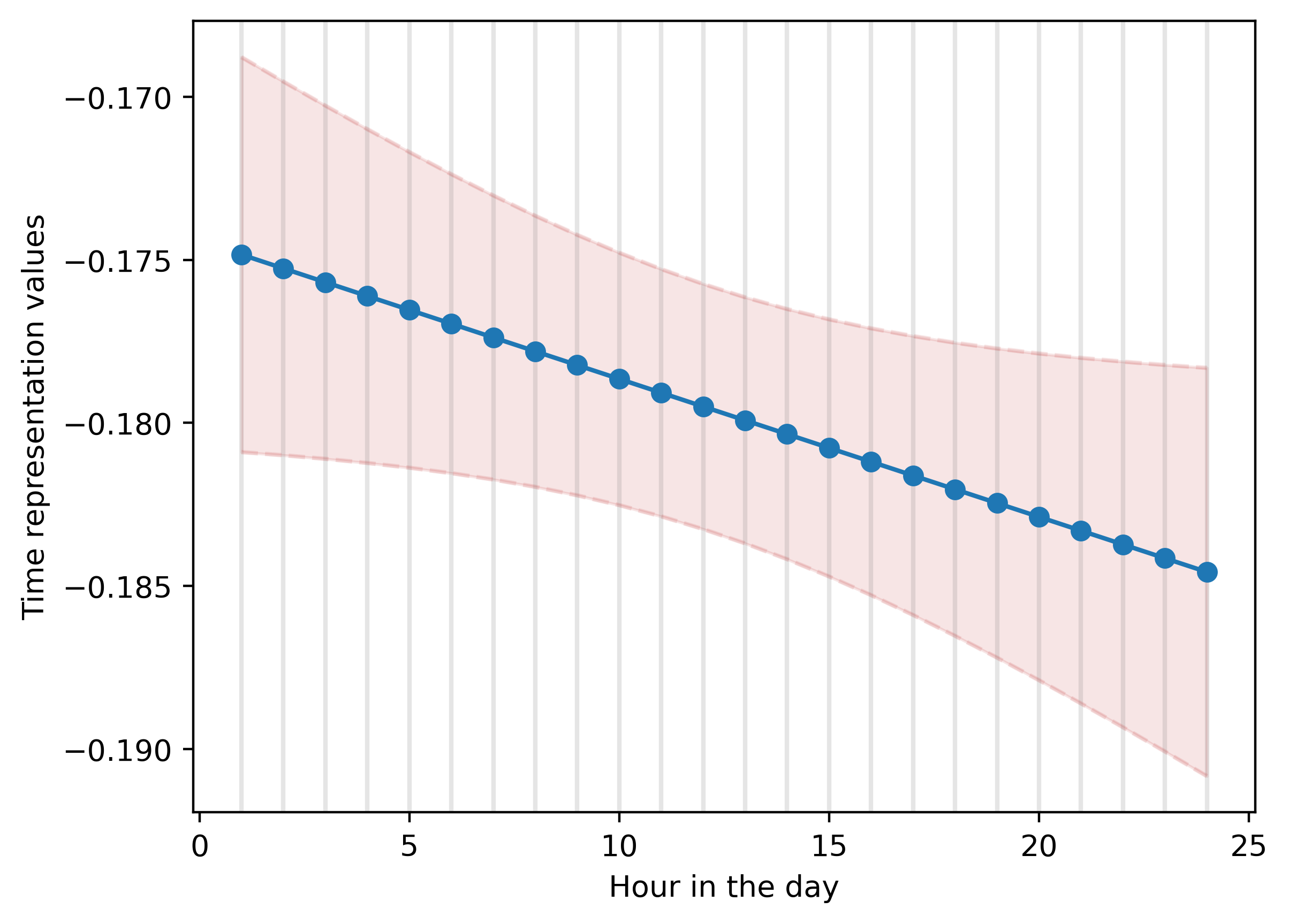}
        \caption{Group 5}
        \label{fig:grp5_init}
    \end{subfigure}
    \hfill
    \begin{subfigure}[b]{0.32\textwidth}
        \centering
        \includegraphics[width=\textwidth]{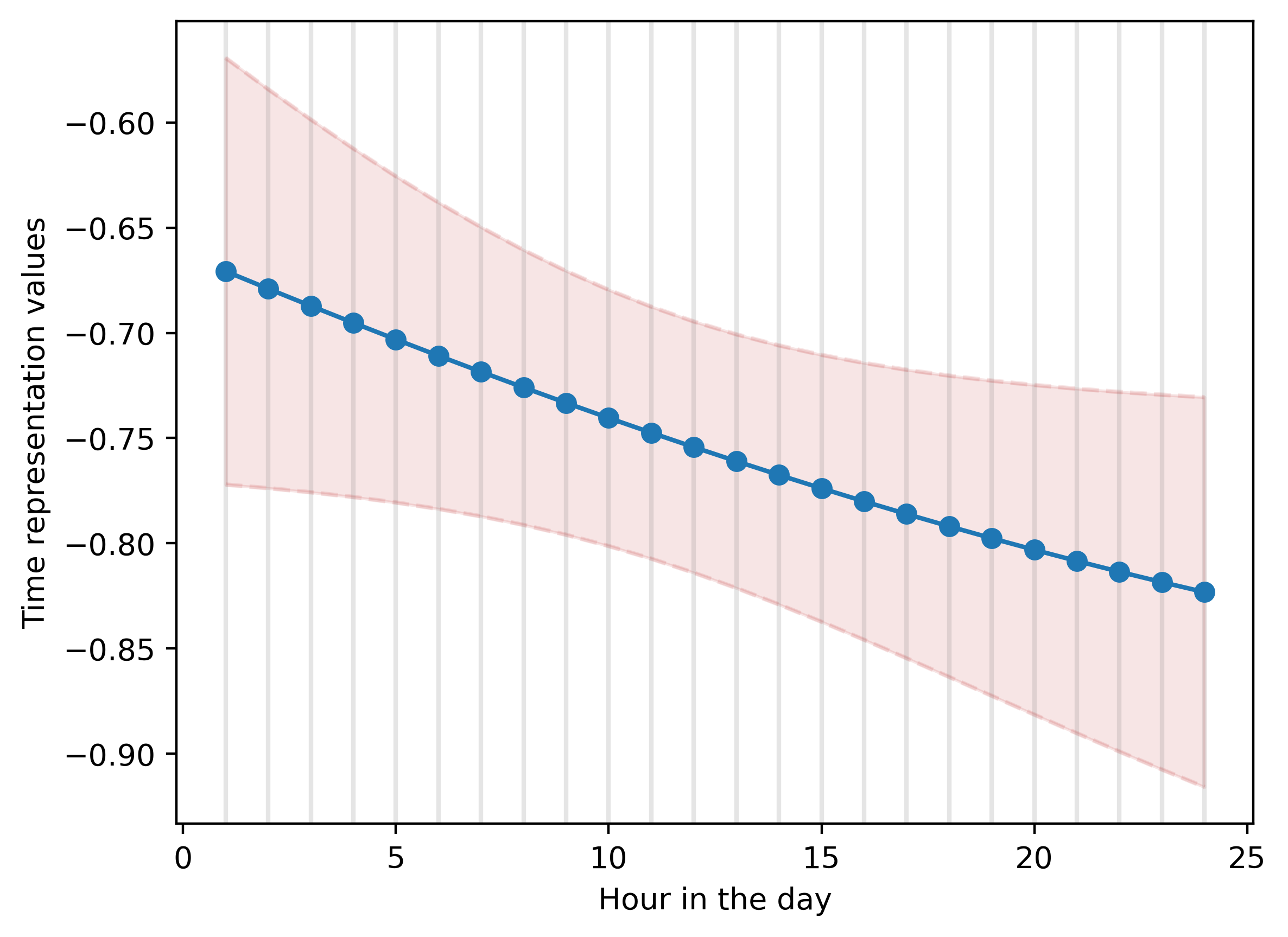}
        \caption{Group 6}
        \label{fig:grp6_init}
    \end{subfigure}
    \caption{Initial values of the groups of the sine time representations.}
    \label{fig:grouped_sines_init}
\end{figure*}

\begin{figure*}[tp]
    \centering
    \begin{subfigure}[b]{0.32\textwidth}
        \centering
        \includegraphics[width=\textwidth]{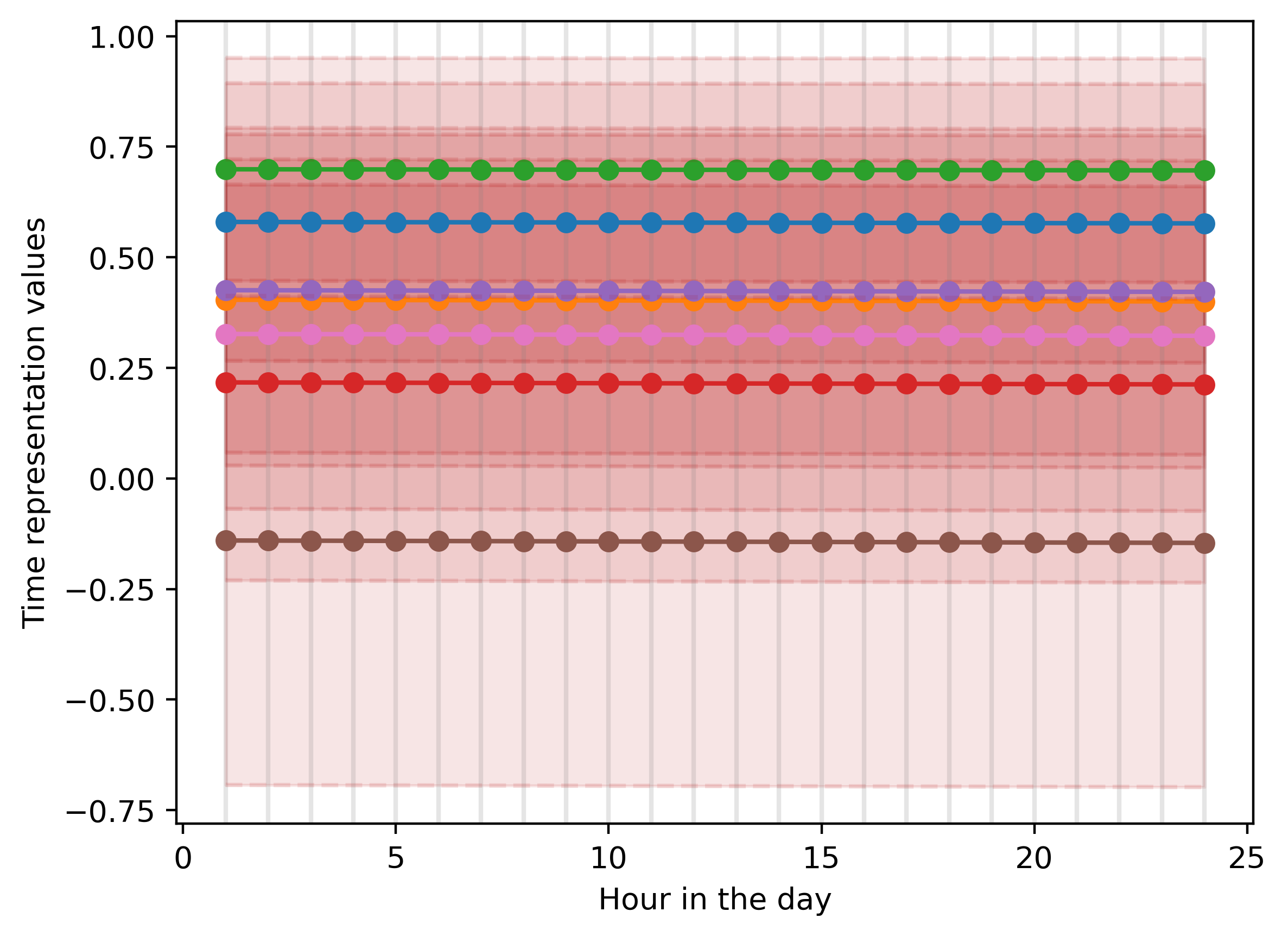}
        \caption{Group 1}
        \label{fig:grp1_time_points_diff}
    \end{subfigure}
    \hfill
    \begin{subfigure}[b]{0.32\textwidth}
        \centering
        \includegraphics[width=\textwidth]{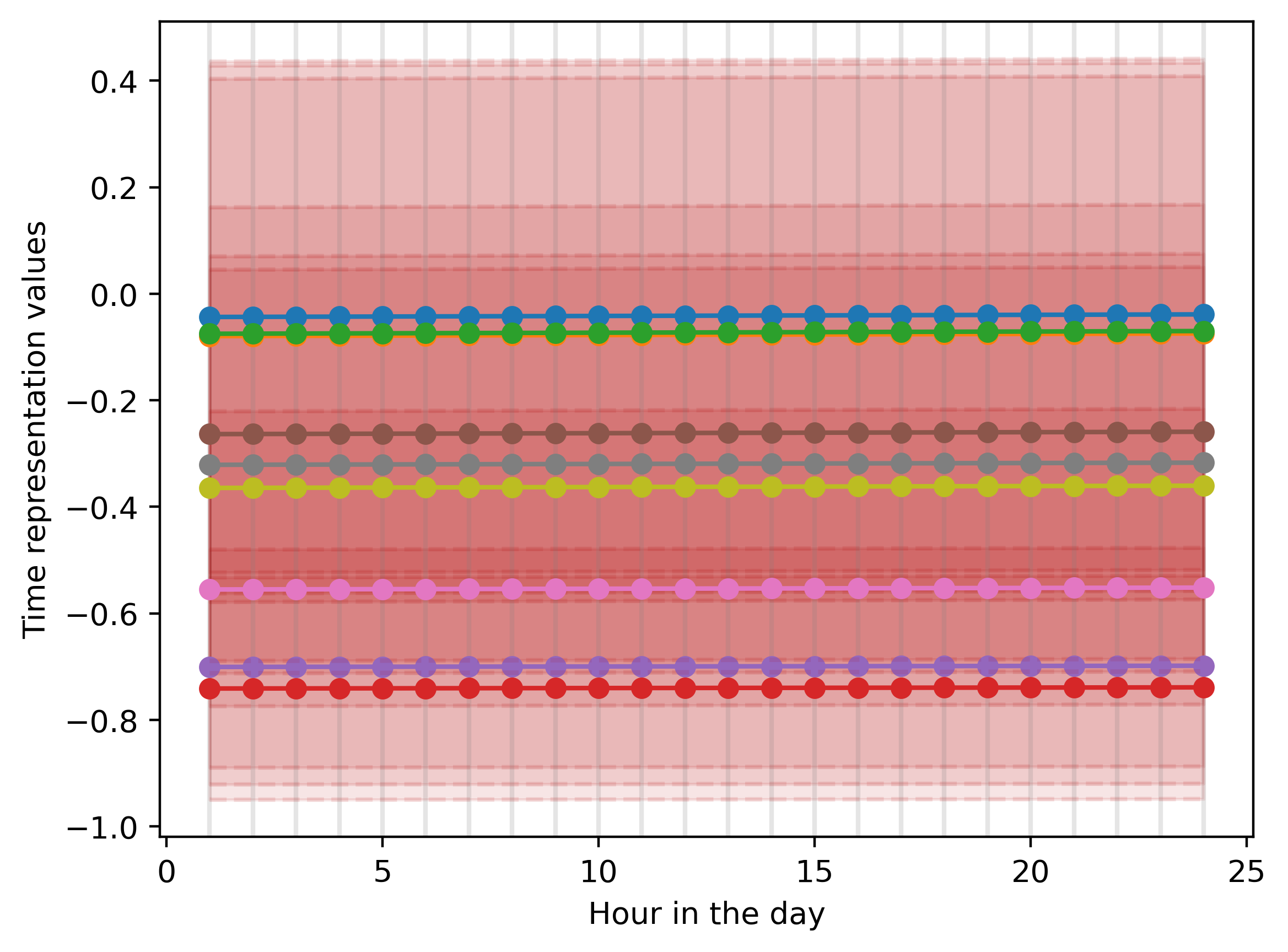}
        \caption{Group 2}
        \label{fig:grp2_time_points_diff}
    \end{subfigure}
    \hfill
    \begin{subfigure}[b]{0.32\textwidth}
        \centering
        \includegraphics[width=\textwidth]{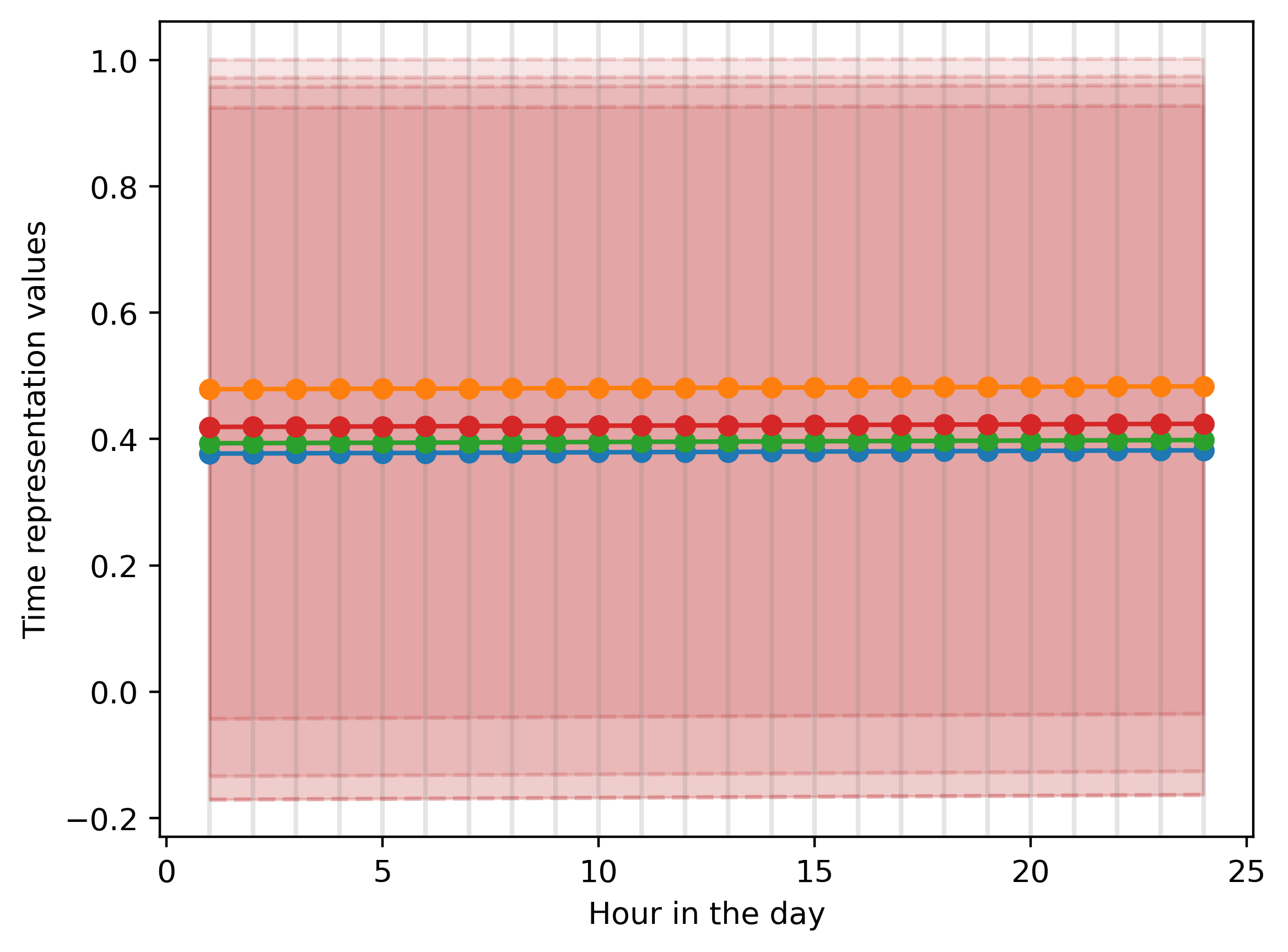}
        \caption{Group 3}
        \label{fig:grp3_time_points_diff}
    \end{subfigure}
    \vfill
    \begin{subfigure}[b]{0.32\textwidth}
        \centering
        \includegraphics[width=\textwidth]{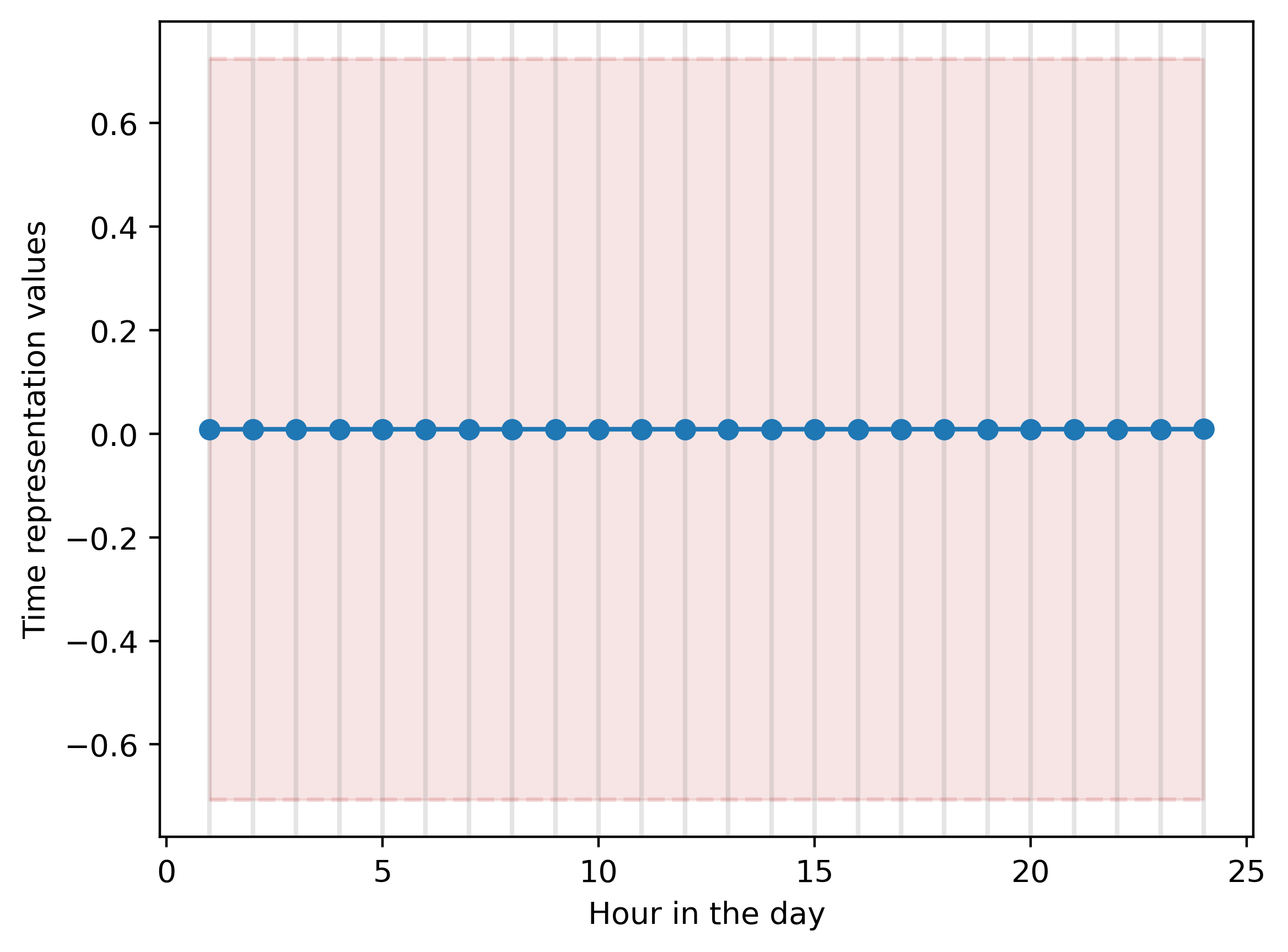}
        \caption{Group 4}
        \label{fig:grp4_time_points_diff}
    \end{subfigure}
    \hfill
    \begin{subfigure}[b]{0.32\textwidth}
        \centering
        \includegraphics[width=\textwidth]{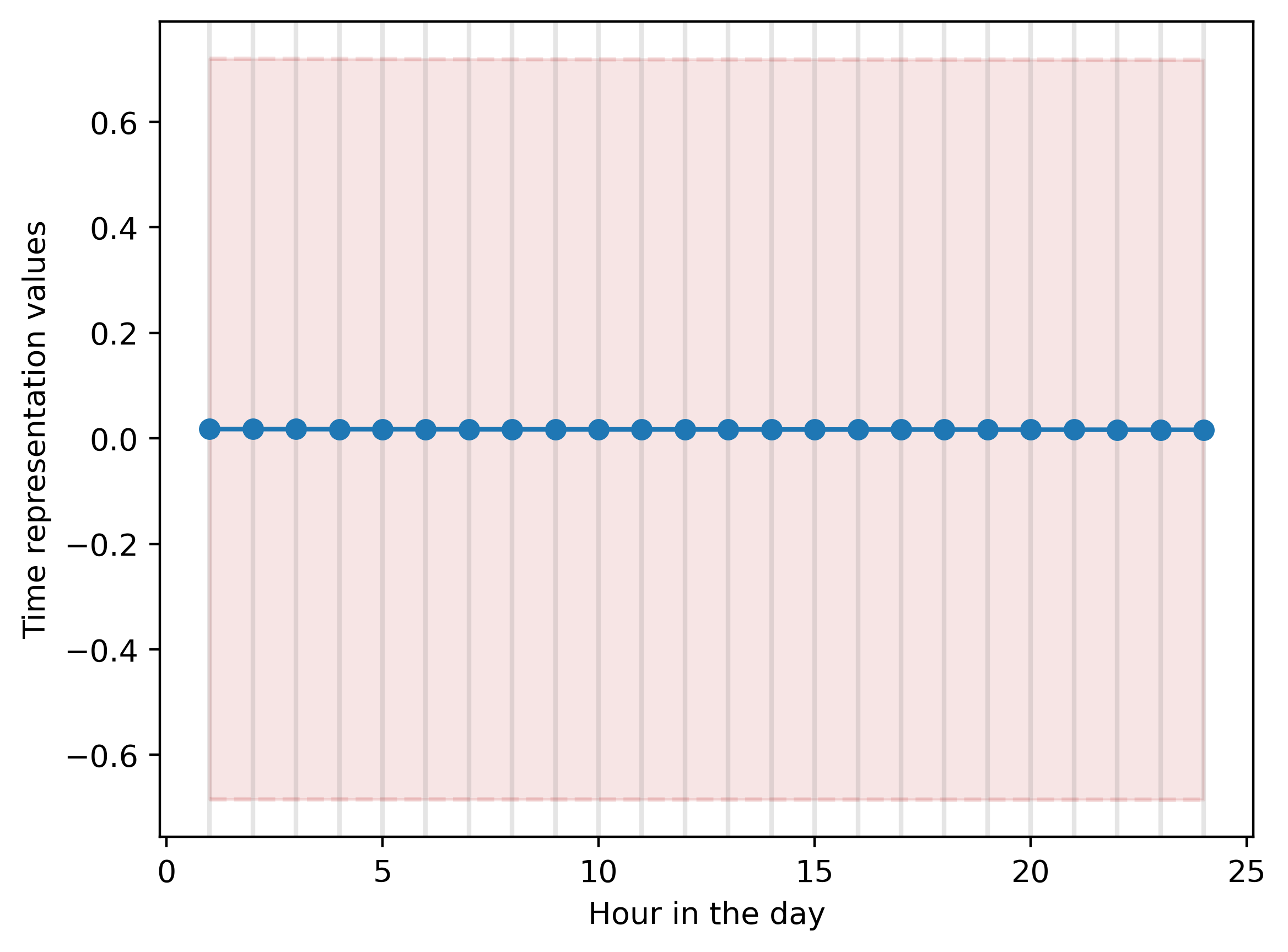}
        \caption{Group 5}
        \label{fig:grp5_time_points_diff}
    \end{subfigure}
    \hfill
    \begin{subfigure}[b]{0.32\textwidth}
        \centering
        \includegraphics[width=\textwidth]{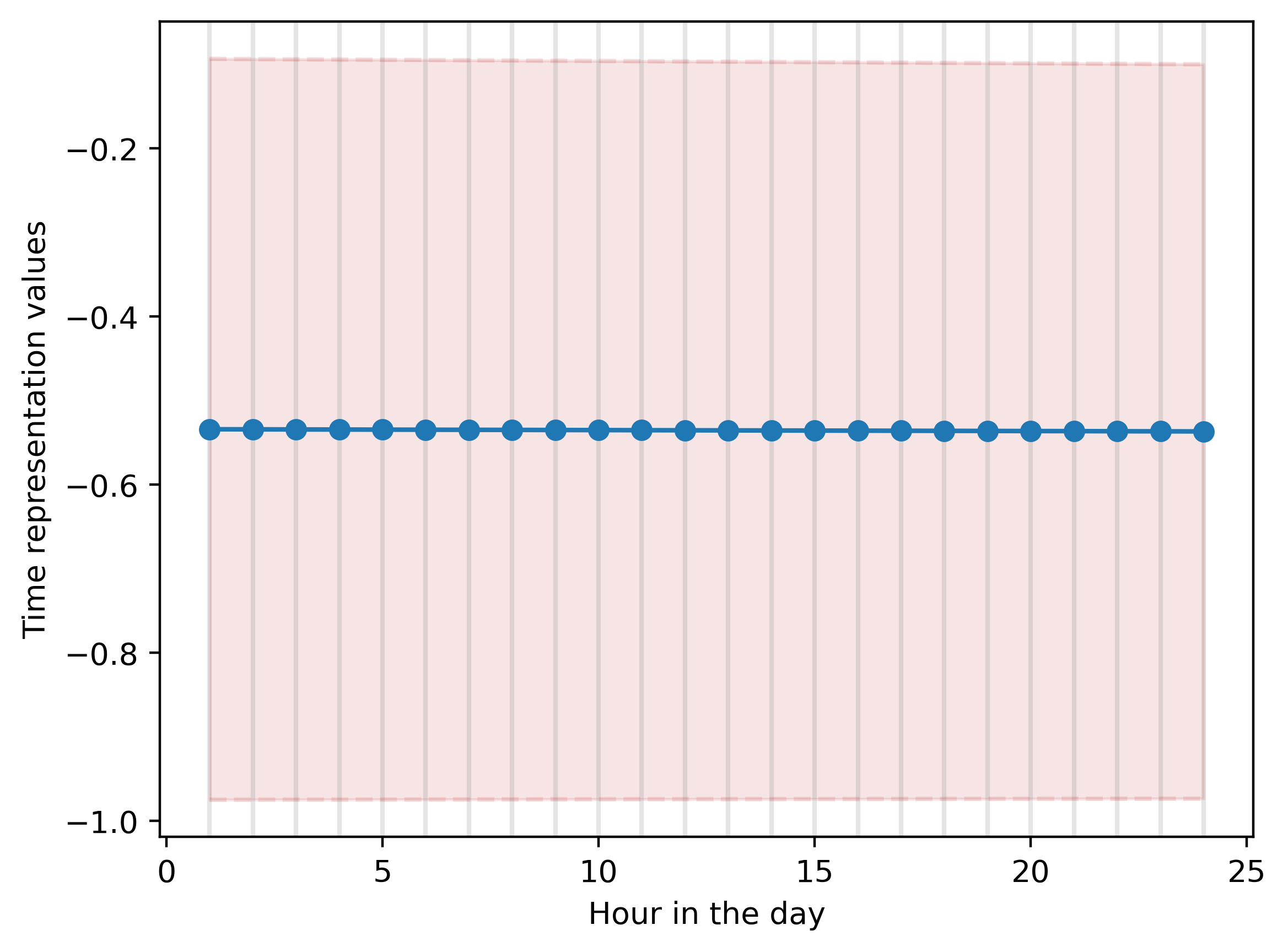}
        \caption{Group 6}
        \label{fig:grp6_time_points_diff}
    \end{subfigure}
    \caption{Learned sine time representations on the keys of the test data. The illustrated representations occurred when the keys and queries had different orders of magnitude.}
    \label{fig:grouped_sines_time_points_diff_scale}
\end{figure*}

\begin{figure*}[tp]
    \centering
    \begin{subfigure}[b]{0.32\textwidth}
        \centering
        \includegraphics[width=\textwidth]{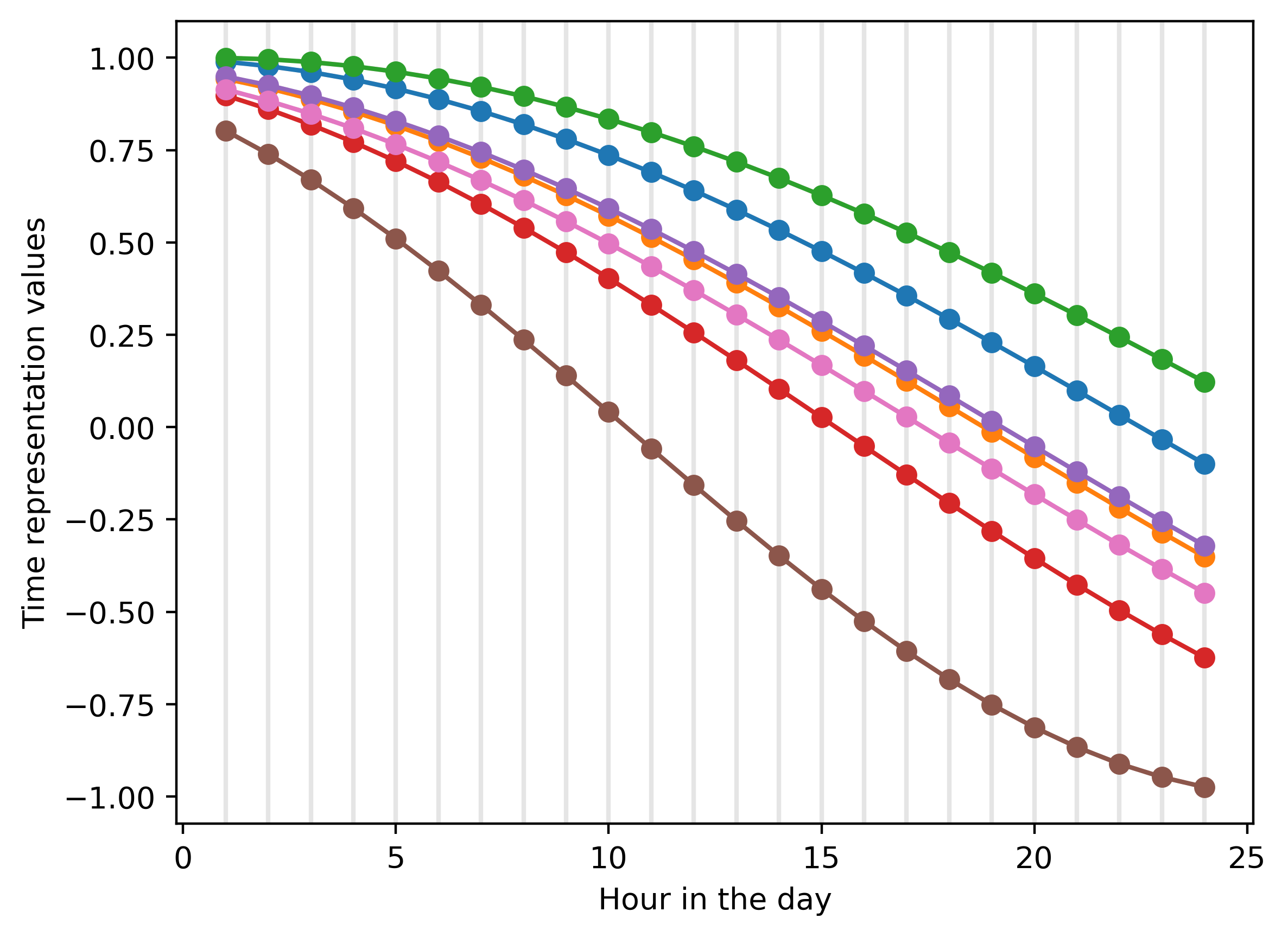}
        \caption{Group 1}
        \label{fig:grp1_lin_diff}
    \end{subfigure}
    \hfill
    \begin{subfigure}[b]{0.32\textwidth}
        \centering
        \includegraphics[width=\textwidth]{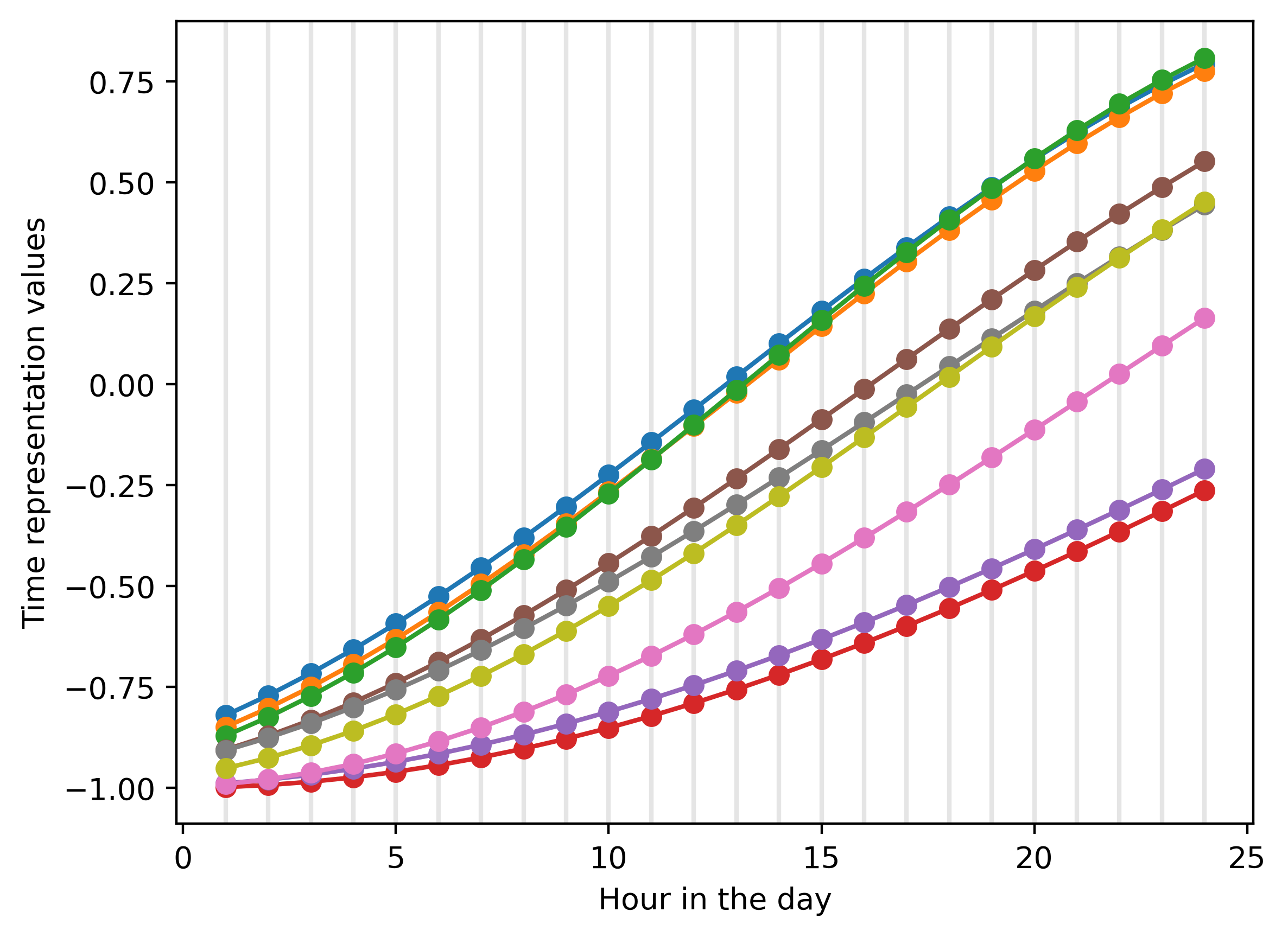}
        \caption{Group 2}
        \label{fig:grp2_lin_diff}
    \end{subfigure}
    \hfill
    \begin{subfigure}[b]{0.32\textwidth}
        \centering
        \includegraphics[width=\textwidth]{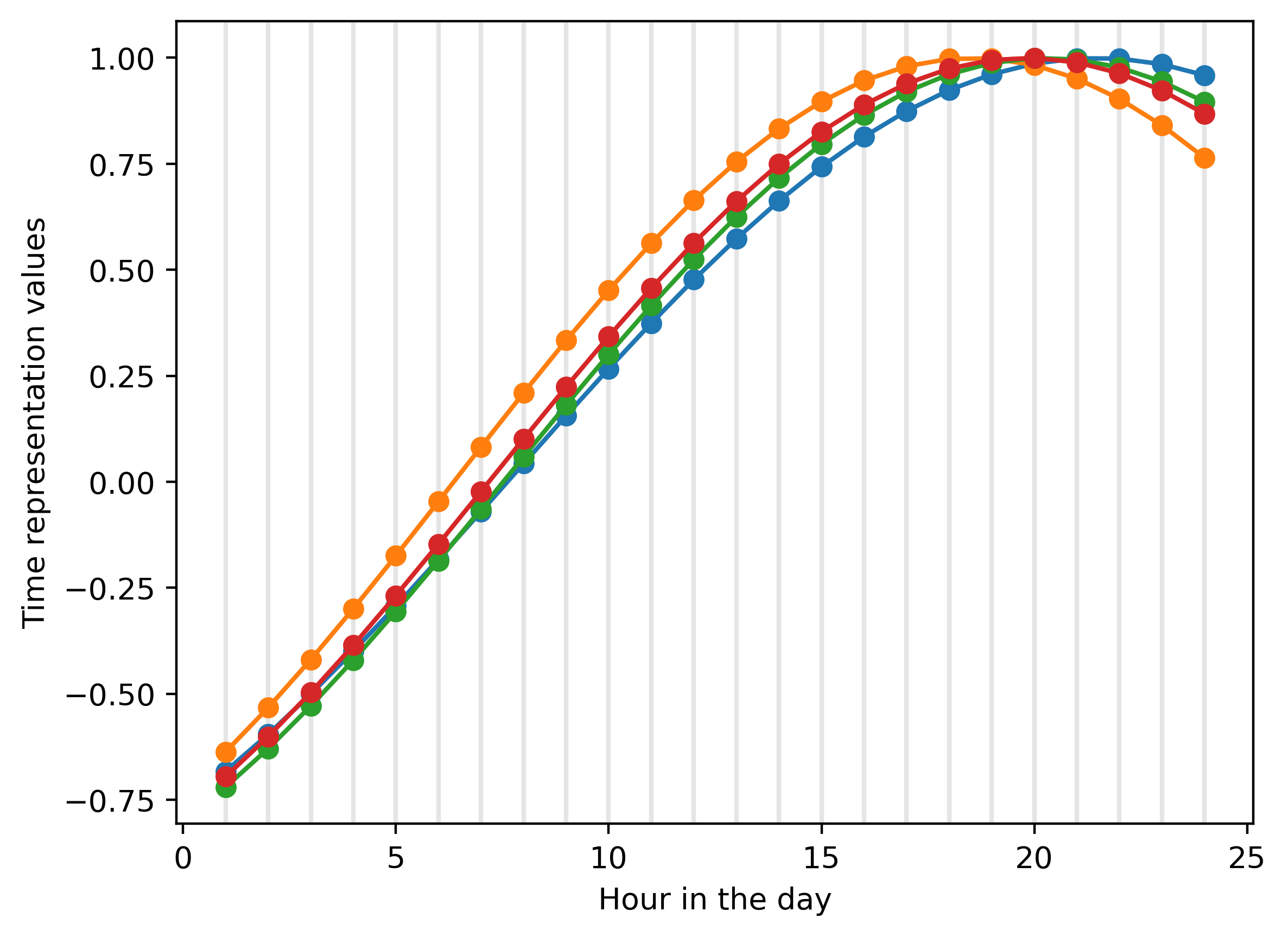}
        \caption{Group 3}
        \label{fig:grp3_lin_diff}
    \end{subfigure}
    \vfill
    \begin{subfigure}[b]{0.32\textwidth}
        \centering
        \includegraphics[width=\textwidth]{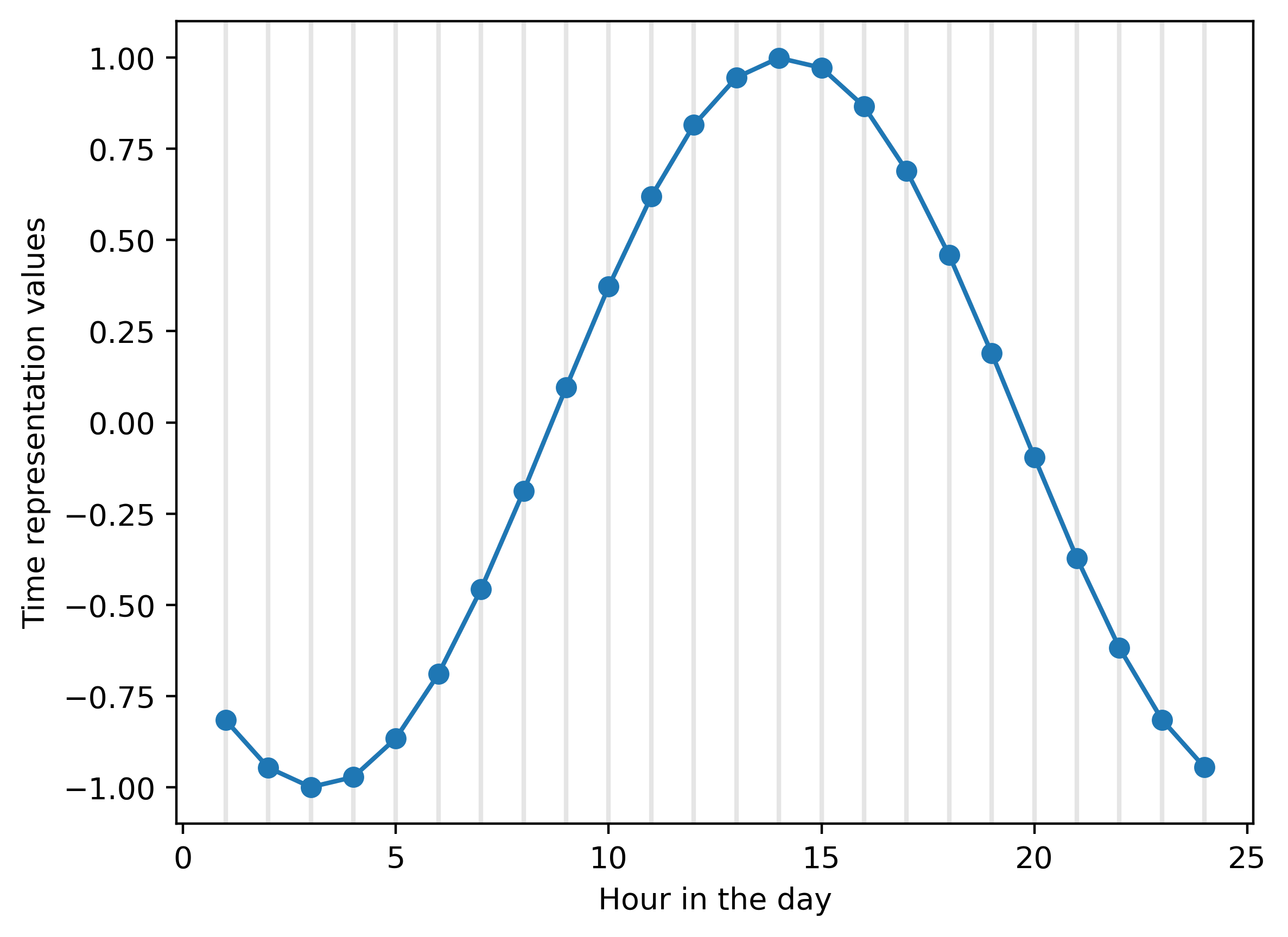}
        \caption{Group 4}
        \label{fig:grp4_lin_diff}
    \end{subfigure}
    \hfill
    \begin{subfigure}[b]{0.32\textwidth}
        \centering
        \includegraphics[width=\textwidth]{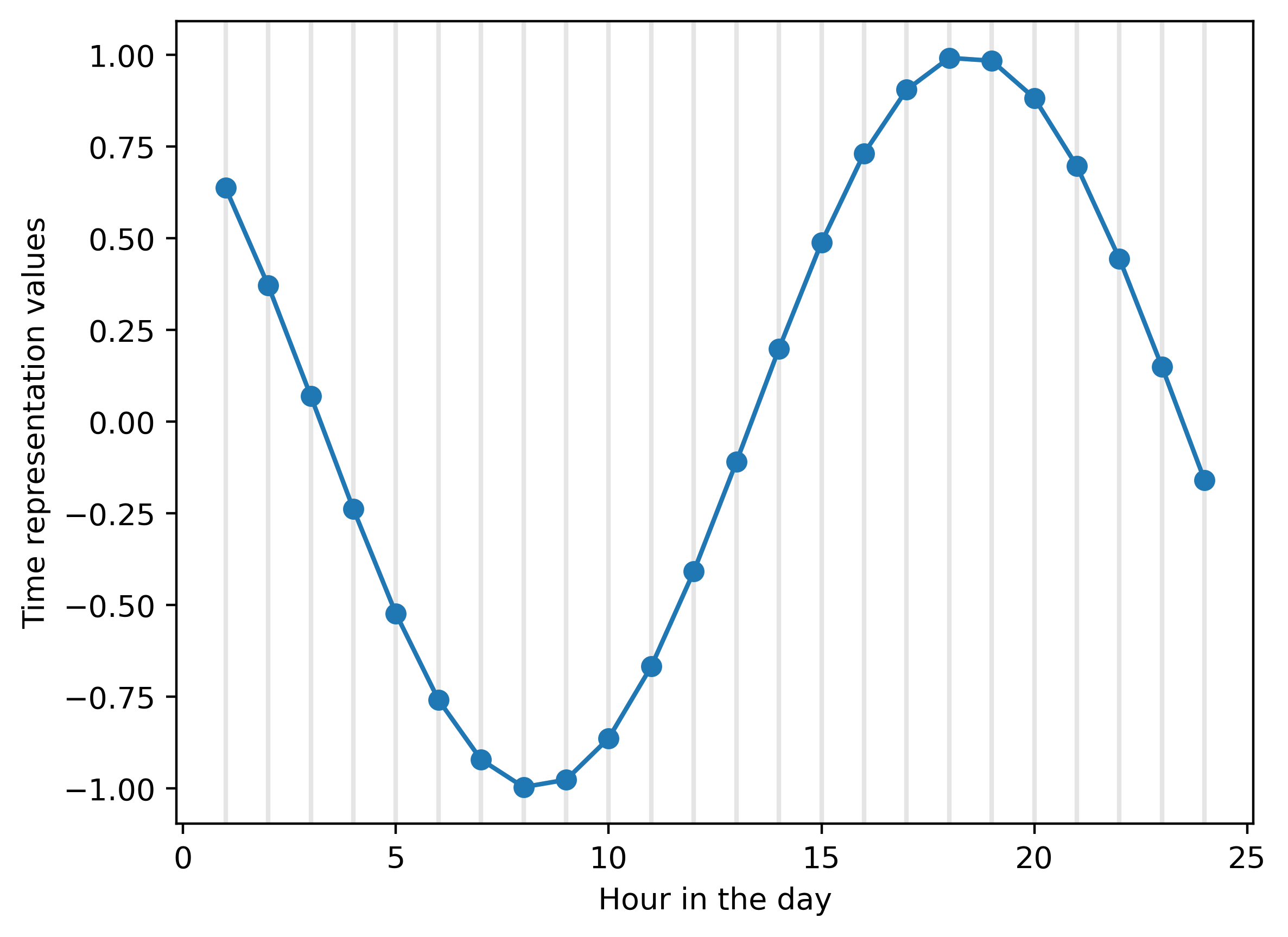}
        \caption{Group 5}
        \label{fig:grp5_lin_diff}
    \end{subfigure}
    \hfill
    \begin{subfigure}[b]{0.32\textwidth}
        \centering
        \includegraphics[width=\textwidth]{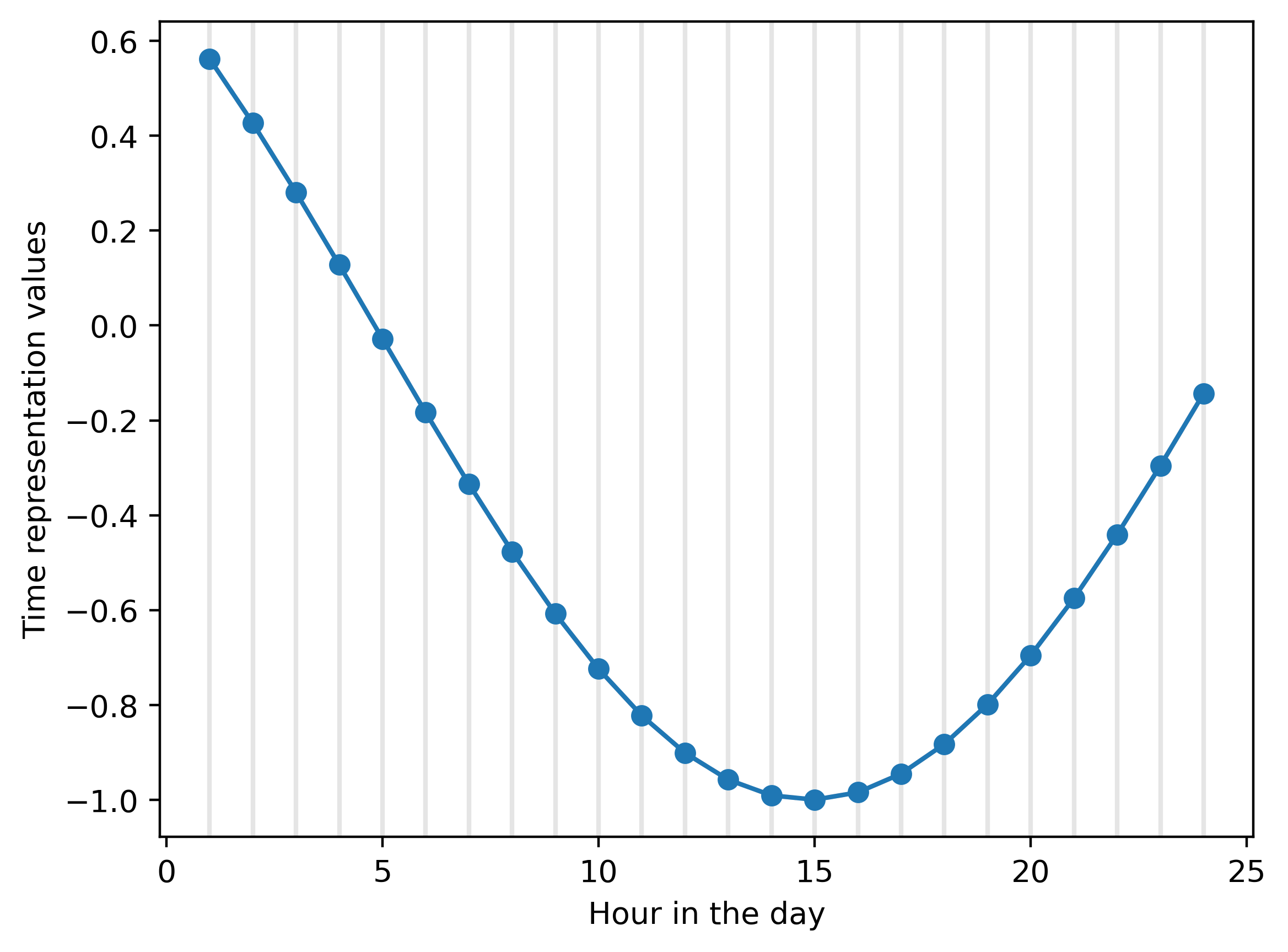}
        \caption{Group 6}
        \label{fig:grp6_lin_diff}
    \end{subfigure}
     \caption{Learned sine time representations on the queries of the test data. The illustrated representations occurred when the keys and queries had different orders of magnitude.}
    \label{fig:grouped_sines_linspace_diff_scale}
\end{figure*}

\begin{figure*}[tp]
    \centering
    \begin{subfigure}[b]{0.32\textwidth}
        \centering
        \includegraphics[width=\textwidth]{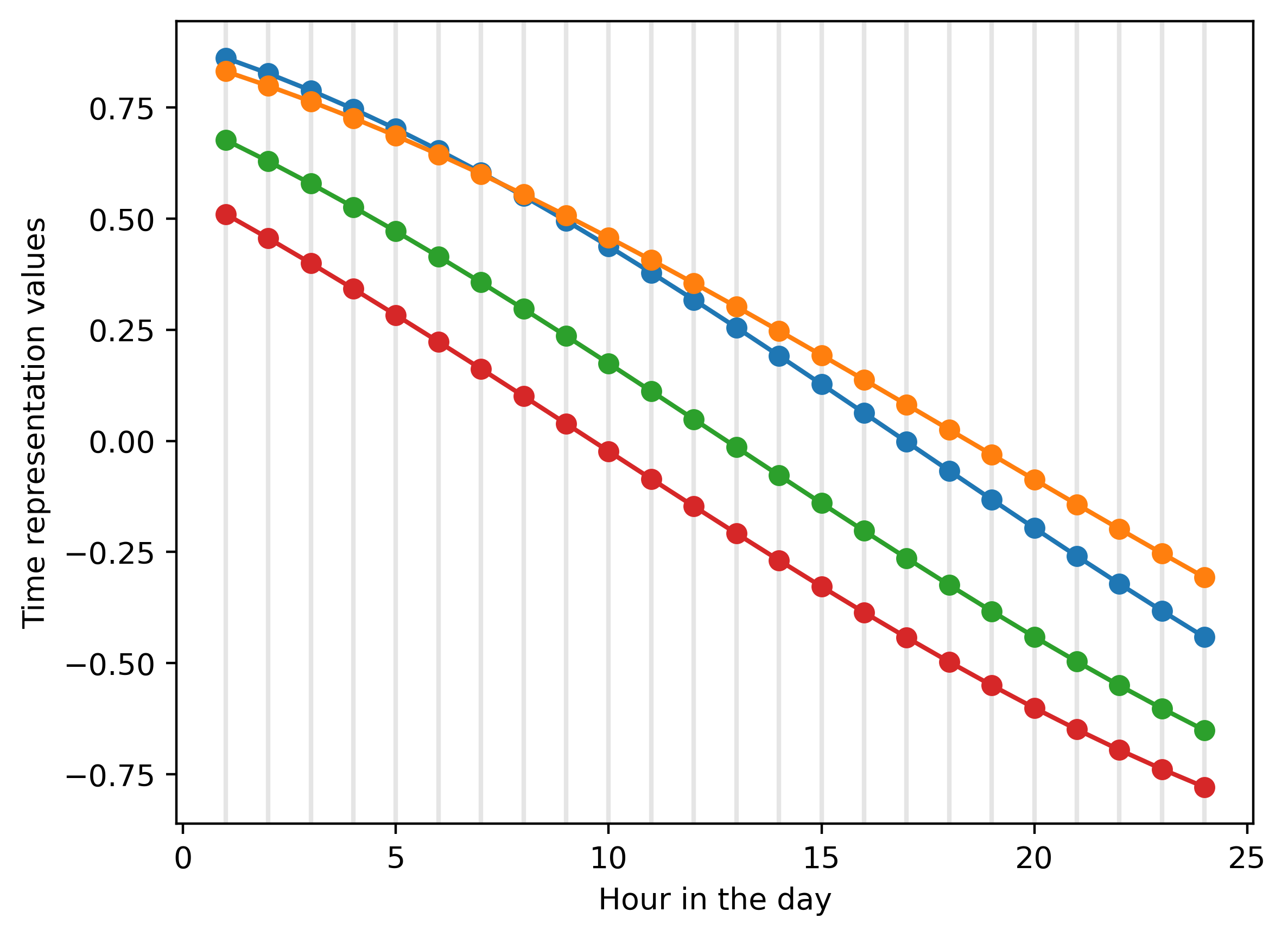}
        \caption{Group 1}
        \label{fig:grp1_same}
    \end{subfigure}
    \hfill
    \begin{subfigure}[b]{0.32\textwidth}
        \centering
        \includegraphics[width=\textwidth]{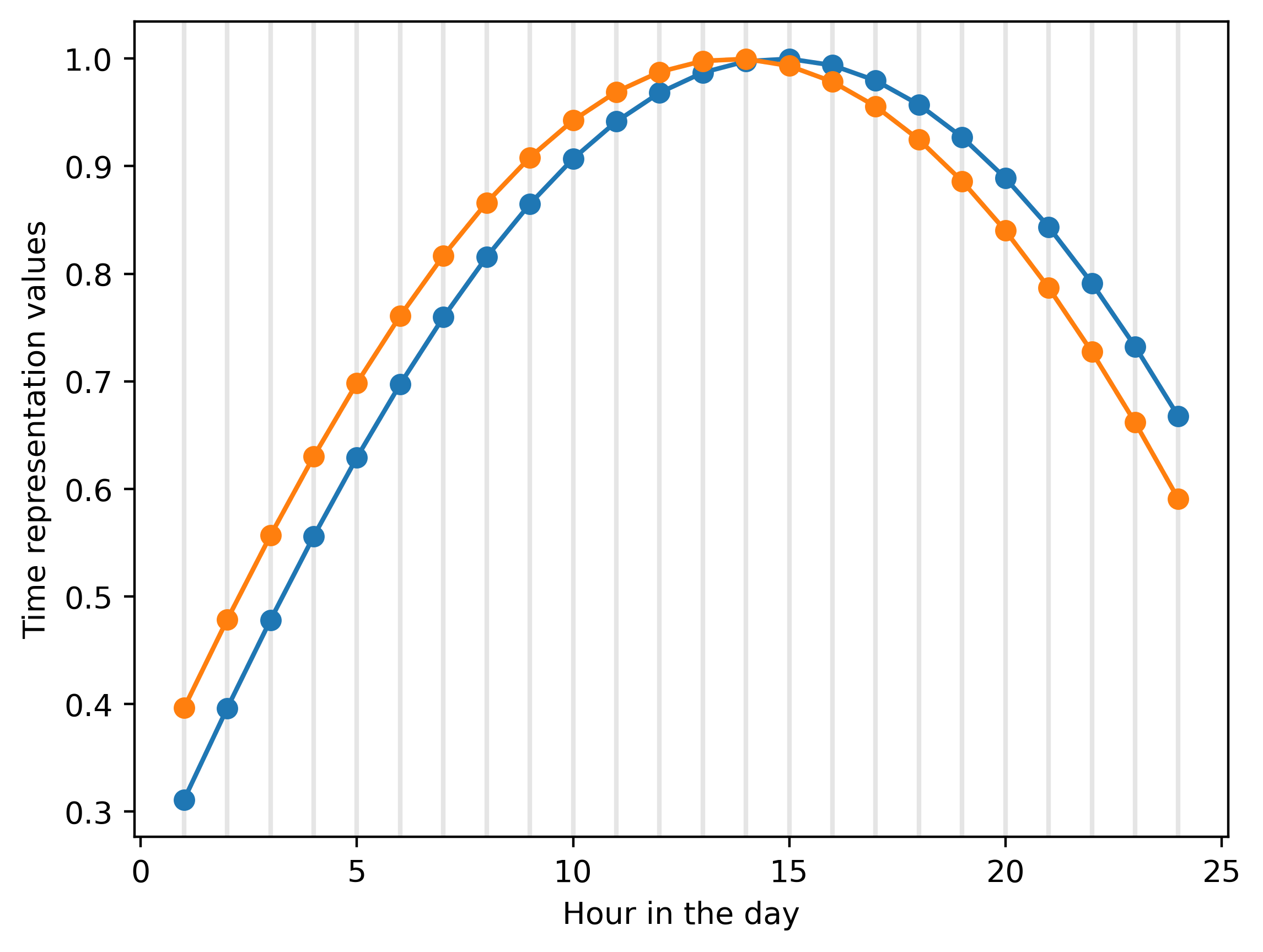}
        \caption{Group 2}
        \label{fig:grp2_same}
    \end{subfigure}
    \hfill
    \begin{subfigure}[b]{0.32\textwidth}
        \centering
        \includegraphics[width=\textwidth]{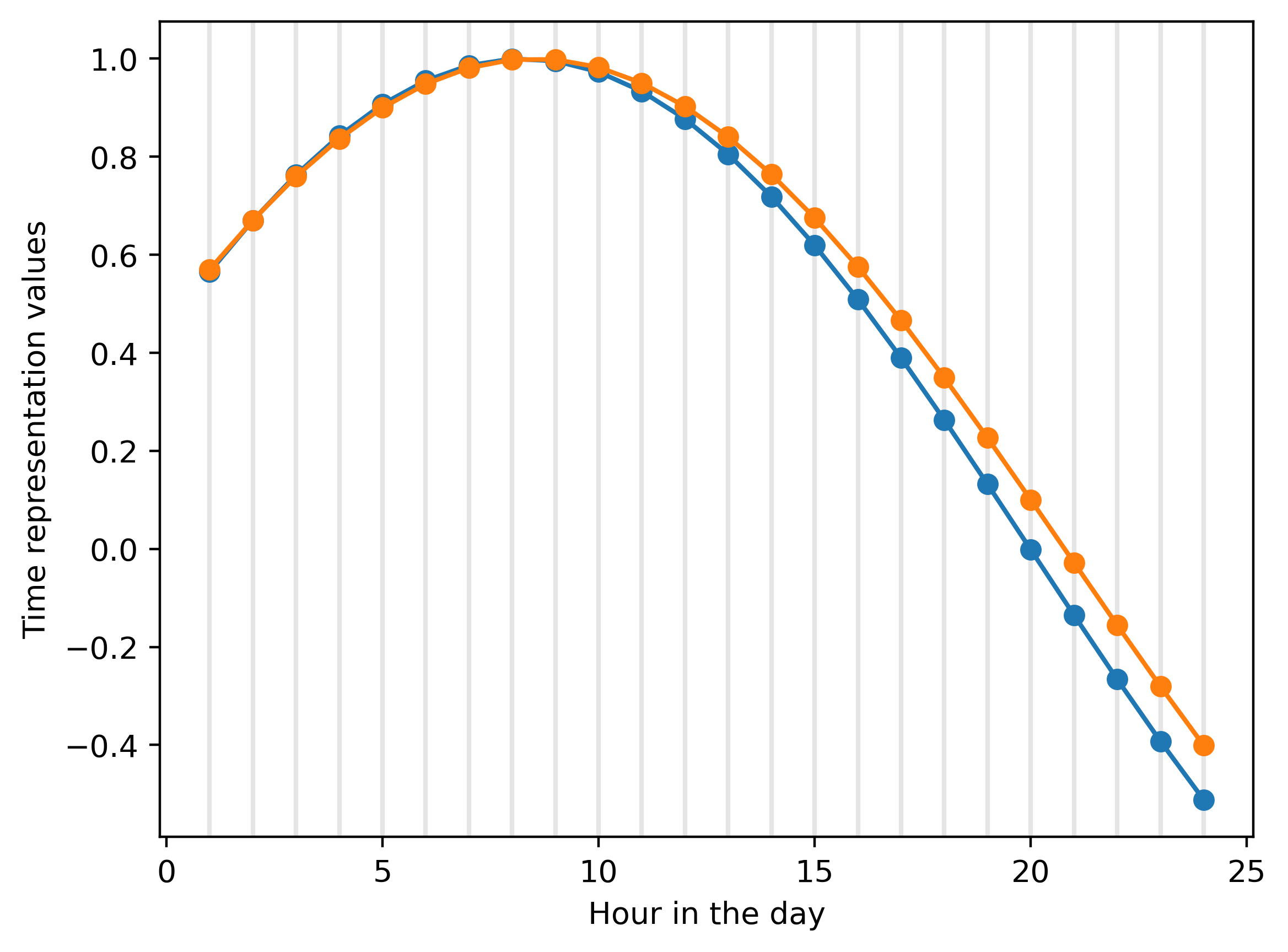}
        \caption{Group 3}
        \label{fig:grp3_same}
    \end{subfigure}
    \vfill
    \begin{subfigure}[b]{0.32\textwidth}
        \centering
        \includegraphics[width=\textwidth]{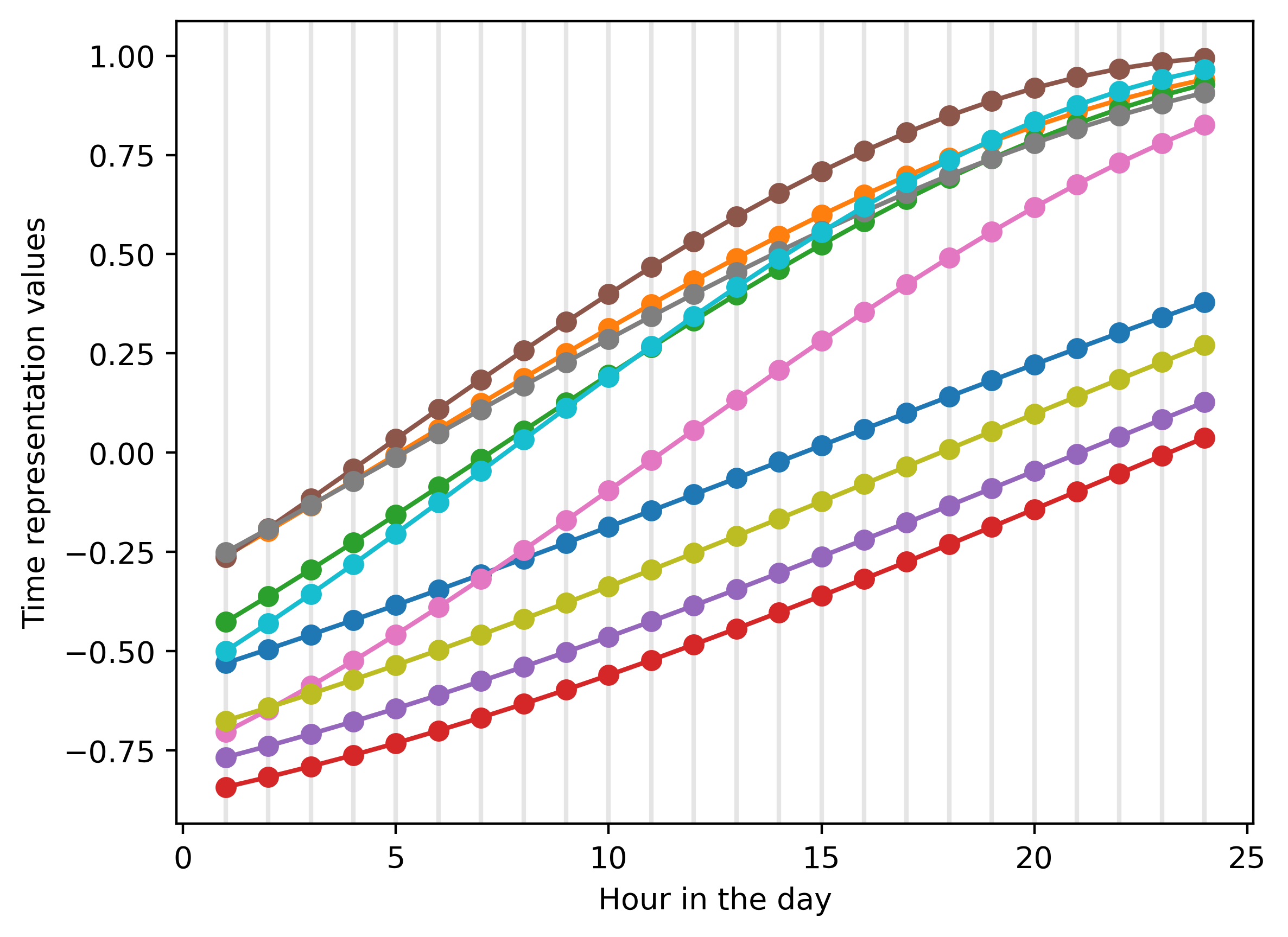}
        \caption{Group 4}
        \label{fig:grp4_same}
    \end{subfigure}
    \hfill
    \begin{subfigure}[b]{0.32\textwidth}
        \centering
        \includegraphics[width=\textwidth]{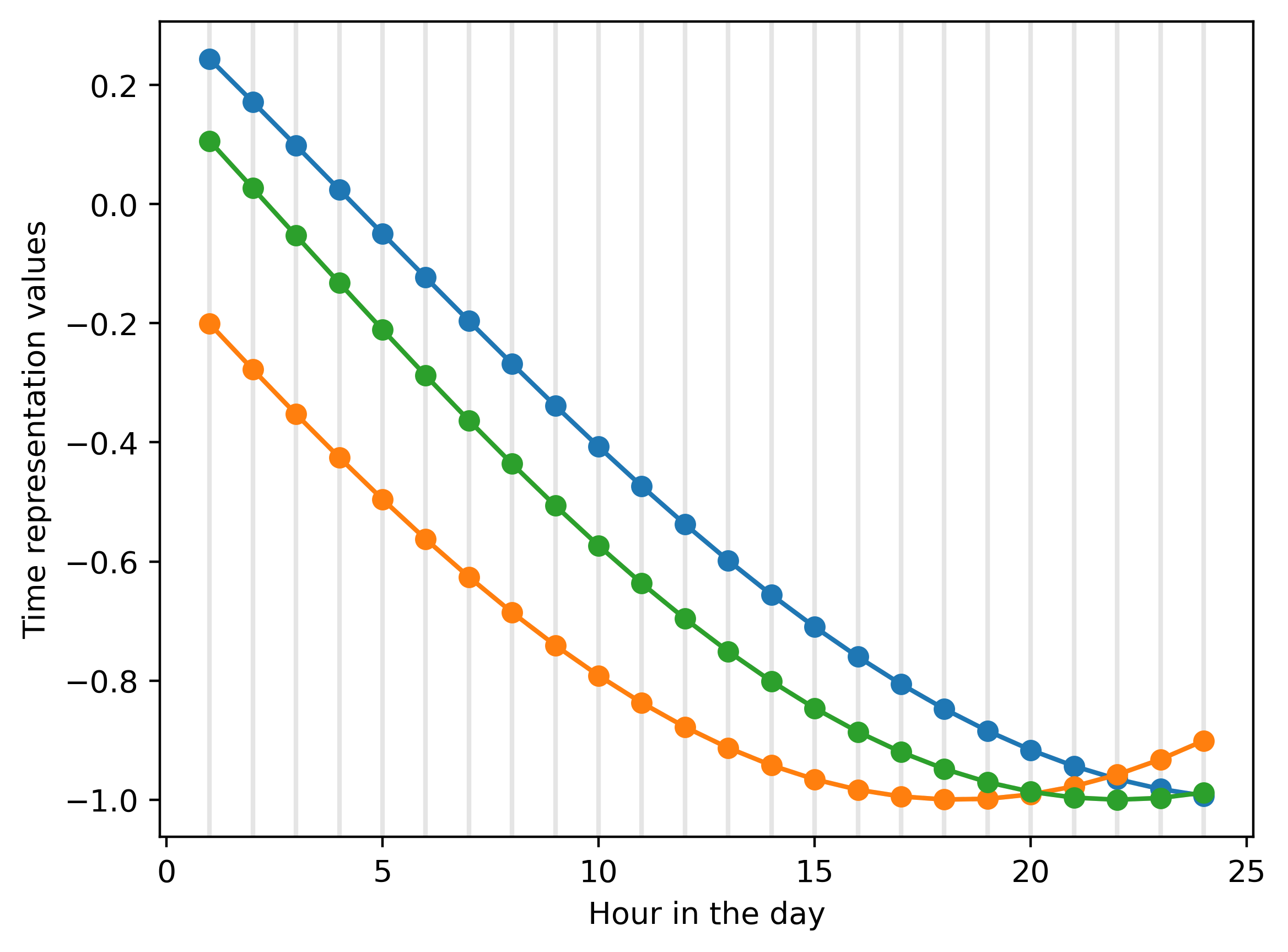}
        \caption{Group 5}
        \label{fig:grp5_same}
    \end{subfigure}
    \hfill
    \begin{subfigure}[b]{0.32\textwidth}
        \centering
        \includegraphics[width=\textwidth]{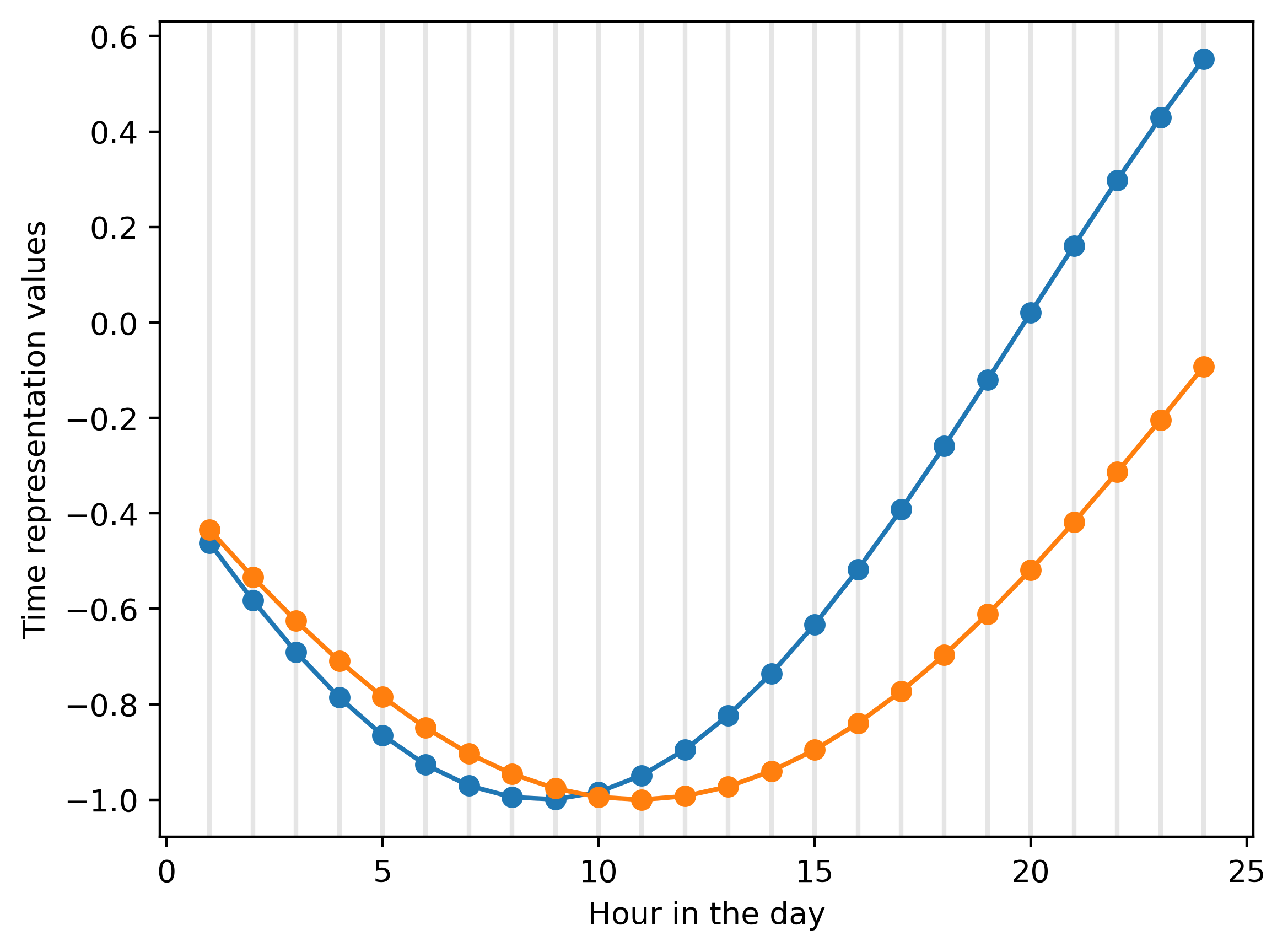}
        \caption{Group 6 }
        \label{fig:grp6_same}
    \end{subfigure}
    \caption{Learned sine time representations on the keys and queries of the test data. The illustrated representations occurred when the keys and queries had the same order of magnitude.}
    \label{fig:grouped_sines_same_scale}
\end{figure*}

As noted in Section~\ref{sec:bg}, we are bound by
the model's architecture to learn \( n-1 \) periodic and \( 1 \)
non-periodic feature representations from our data, with \( n \) being
the number of the final output points of the time series (in our case, 24).

When subjecting the periodic time representation to be a triangular pulse, the model starts from very narrow pulses and then proceeds to learn a combination of those, as presented previously in Figure \ref{fig:learned_pulse}.
On the other hand, when we define the periodicity as a sine function, the initial weight initialization (which sets the weights to very small values) makes
the initial representations a very short part of the sinusoidal curve
(so short that visually appears as linear in our graphs).
The model then learns to scale the time vectors to model more meaningful time representations.
An example of this behavior is presented in Figure~\ref{fig:sine_b4_and_after}.

However, one might wonder: are all of the 23 learned time representations meaningful?
Since it is hard to track the influence of every representation on the classification task, we can only speculate.
In Figure \ref{fig:grouped_sines} are demonstrated all the periodic sine features, grouped by their similarity.
Most time features fall into the first three groups. 
The last three groups, which only include one representation each, seem to be the most explainable regarding our knowledge of the domain. 
They capture larger periods that lead to the numerical correlation of different hours in a day.
We believe that these are the most important features on which the model's prediction relies and also that when it started to learn these features, it neglected the others and assigned smaller weights to them.
To substantiate this assertion we illustrate in Figure \ref{fig:grouped_sines_init} the corresponding groups at the start of the learning process.
It can be claimed that the first three groups exhibit minimal change, whereas the last three change visibly. 

To model the time points for each example, we normalized each timestamp using the Unix timestamps from our data's date range.
The resulting time points range from 0 to 1, but within a daily sample, the hourly time points vary slightly.
The Transformer's architecture utilizes the key/value/query formulation \cite{og_attention}.
The query is the information that is being looked for, the key is the context or reference, and the value is the content that is being searched.
In the context of \term{mTAN}, the initial keys are the time points and the initial queries are vectors with values equally distributed from 0 to 1.
Both of these terms are then passed through the same learnable time layers.
Since, initially, the keys and the queries have a different order of magnitude it is expected that either the model will increase the magnitude of the keys or decrease the magnitude of the queries.
Through our experiments we discovered that the former occurs.
As presented in Figure~\ref{fig:grouped_sines_time_points_diff_scale} the smaller key values correspond to a smaller part of the sine wave, whereas the larger query values cover a larger part of the sine wave (Figure~\ref{fig:grouped_sines_linspace_diff_scale}).

To further evaluate our claim, we modified the time point creation so that both queries and keys have the same values.
To do so, we assigned each time point referring to an hour in a day its normalized index by dividing it with the total number of hours in a day.
As expected, the learned representation then was the same for both keys and queries, as illustrated in Figure \ref{fig:grouped_sines_same_scale}.
Comparing them to the previously learned representations in Figures~\ref{fig:grouped_sines_time_points_diff_scale} and~\ref{fig:grouped_sines_linspace_diff_scale}, the evaluation of the keys has indeed scaled up, while the evaluation of the queries has not changed drastically.

The main reason we opted for normalizing the time steps using UNIX time was that it slightly performed better for the classification task.
We believe that this is the case because when normalizing the time points considering only the hour, we lose any other characteristic that relates to the progression of time.
This further corroborates that (as we extensively stressed earlier)
the non-periodical features are crucial for our application.

\section{Conclusions}\label{sec:conc}

We studied the impact of time embeddings when applying Transformer
models to timeseries analysis, and specifically the impact of using a
very informed, application-specific time representation, a generic
sinusoidal representation known to capture periodic phenomena well,
and a method for learning the parameters that best fit either of these
two to the specific dataset at hand.

Our experiments concluded that Transformers (and DNNs in general we
could argue) are not very amenable to over-engineering the time
representation due to side-effects that are difficult to mitigate.
This can be seen as a negative quality in applications such as ours
where there is extensive prior knowledge on a well-studied phenomenon,
but can also be construed as a positive quality as the network was
able to \quotes{discover} the knowledge we were trying to convey.
Especially the comparison between the sinusoidal prior and learned
representations is very promising in this respect, at the
maximally-flexible \term{mTAN} network recovered almost all the accuracy
of the prior-time sinusoidal network.

A second level of analysis delved into the nature of the features
that \term{mTAN} learned. This analysis has demonstrated the ability of
the linear layer - sinusoidal activation function architecture to
very closely approximate the clearly non-smooth behavior
discussed in Section~\ref{sec:time} and shown in
Figure~\ref{fig:mean_pyranometer_approx}.

It should, however, be noted that our analysis was restricted to what
we could indirectly observe by trying out different parameterizations
and speculating based on our understanding of how the network is
trained. Besides any general advancements in methodologies for
explaining neural networks, this also showed us a path for future
work specifically targeting \term{mTAN}. Since one of the hindrances was that
we were unable to observe the effect of each feature due the
distributed nature of the classifier, we would have liked to
systematically explore the effect of incrementally increasing the
number of features and observing which features gets learned
(interpreted as, is the most impactful for reducing loss), which is
learned second and so on. However, the linear algebra behind \term{mTAN}
ties the dimensionality of the feature representation to the
dimensionality of the input, as we cannot de-couple the encoding
of periodicity and the encoding of linear time. Since (as argued
above) linear time is critical for performance, we are stuck with a
pre-defined dimensionality for the sine features as well.

Our envisaged future research is to re-work the linear algebra of the
\term{mTAN} so that we can de-couple the encoding of the periodic and linear
time representation, allowing to have only the latter be constrained by
the dimensionality of the input.
In the specific experiment presented here, this would translate to a
better understanding of what happens when we inform the Transformer of
the bias we want to apply regarding periodicities. Which, in its turn,
is expected to lead to methods for affording the human operator
intuitive and effective control of the network.

\begin{acks}
This research has been co-financed by the European Union and
Greek national funds through the program
\quotes{Flagship actions in interdisciplinary scientific areas with
a special interest in the connection with the production network}
--- GREEN SHIPPING --- TAEDR-0534767 (Acronym: NAVGREEN).
For more information please visit \url{https://navgreen.gr}

This research was co-funded by the European Union under GA
no.~101135782 (MANOLO project). Views and opinions expressed are
however those of the authors only and do not necessarily reflect those
of the European Union or CNECT. Neither the European Union nor CNECT
can be held responsible for them.

AWS resources were provided by the National Infrastructures for
Research and Technology GRNET and funded by the EU Recovery and
Resiliency Facility.
\end{acks}


\begin{thebibliography}{7}

\ifx \showCODEN    \undefined \def \showCODEN     #1{\unskip}     \fi
\ifx \showDOI      \undefined \def \showDOI       #1{#1}\fi
\ifx \showISBNx    \undefined \def \showISBNx     #1{\unskip}     \fi
\ifx \showISBNxiii \undefined \def \showISBNxiii  #1{\unskip}     \fi
\ifx \showISSN     \undefined \def \showISSN      #1{\unskip}     \fi
\ifx \showLCCN     \undefined \def \showLCCN      #1{\unskip}     \fi
\ifx \shownote     \undefined \def \shownote      #1{#1}          \fi
\ifx \showarticletitle \undefined \def \showarticletitle #1{#1}   \fi
\ifx \showURL      \undefined \def \showURL       {\relax}        \fi
\providecommand\bibfield[2]{#2}
\providecommand\bibinfo[2]{#2}
\providecommand\natexlab[1]{#1}
\providecommand\showeprint[2][]{arXiv:#2}

\bibitem[Che et~al\mbox{.}(2018)]%
        {che_etal}
\bibfield{author}{\bibinfo{person}{Zhengping Che}, \bibinfo{person}{Sanjay
  Purushotham}, \bibinfo{person}{Kyunghyun Cho}, \bibinfo{person}{David
  Sontag}, {and} \bibinfo{person}{Yan Liu}.} \bibinfo{year}{2018}\natexlab{}.
\newblock \showarticletitle{Recurrent Neural Networks for Multivariate Time
  Series with Missing Values}.
\newblock \bibinfo{journal}{\emph{Scientific Reports}}  \bibinfo{volume}{8}
  (\bibinfo{date}{04} \bibinfo{year}{2018}).
\newblock
Issue 1.


\bibitem[Neil et~al\mbox{.}(2016)]%
        {neil_etal}
\bibfield{author}{\bibinfo{person}{Daniel Neil}, \bibinfo{person}{Michael
  Pfeiffer}, {and} \bibinfo{person}{Shih-Chii Liu}.}
  \bibinfo{year}{2016}\natexlab{}.
\newblock \showarticletitle{Phased LSTM: Accelerating Recurrent Network
  Training for Long or Event-based Sequences}. In
  \bibinfo{booktitle}{\emph{Advances In Neural Information Processing
  Systems}}. \bibinfo{pages}{3882--3890}.
\newblock


\bibitem[Pham et~al\mbox{.}(2017)]%
        {pham_etal}
\bibfield{author}{\bibinfo{person}{Trang Pham}, \bibinfo{person}{Truyen Tran},
  \bibinfo{person}{Dinh Phung}, {and} \bibinfo{person}{Svetha Venkatesh}.}
  \bibinfo{year}{2017}\natexlab{}.
\newblock \showarticletitle{Predicting healthcare trajectories from medical
  records: A deep learning approach}.
\newblock \bibinfo{journal}{\emph{Journal of Biomedical Informatics}}
  \bibinfo{volume}{69} (\bibinfo{year}{2017}), \bibinfo{pages}{218--229}.
\newblock
\showISSN{1532-0464}


\bibitem[Schirmer et~al\mbox{.}(2022)]%
        {mona_etal}
\bibfield{author}{\bibinfo{person}{Mona Schirmer}, \bibinfo{person}{Mazin
  Eltayeb}, \bibinfo{person}{Stefan Lessmann}, {and} \bibinfo{person}{Maja
  Rudolph}.} \bibinfo{year}{2022}\natexlab{}.
\newblock \showarticletitle{Modeling Irregular Time Series with Continuous
  Recurrent Units}. In \bibinfo{booktitle}{\emph{Proceedings of the 39th
  International Conference on Machine Learning, 17-23 Jul 2022}},
  Vol.~\bibinfo{volume}{162}.
\newblock


\bibitem[Shukla and Marlin(2021)]%
        {mTAN}
\bibfield{author}{\bibinfo{person}{Satya~Narayan Shukla} {and}
  \bibinfo{person}{Benjamin Marlin}.} \bibinfo{year}{2021}\natexlab{}.
\newblock \showarticletitle{Multi-Time Attention Networks for Irregularly
  Sampled Time Series}. In \bibinfo{booktitle}{\emph{International Conference
  on Learning Representations}}.
\newblock


\bibitem[Vaswani et~al\mbox{.}(2017)]%
        {og_attention}
\bibfield{author}{\bibinfo{person}{Ashish Vaswani}, \bibinfo{person}{Noam
  Shazeer}, \bibinfo{person}{Niki Parmar}, \bibinfo{person}{Jakob Uszkoreit},
  \bibinfo{person}{Llion Jones}, \bibinfo{person}{Aidan~N Gomez},
  \bibinfo{person}{\L~ukasz Kaiser}, {and} \bibinfo{person}{Illia Polosukhin}.}
  \bibinfo{year}{2017}\natexlab{}.
\newblock \showarticletitle{Attention is All you Need}. In
  \bibinfo{booktitle}{\emph{Advances in Neural Information Processing
  Systems}}, \bibfield{editor}{\bibinfo{person}{I.~Guyon},
  \bibinfo{person}{U.~Von Luxburg}, \bibinfo{person}{S.~Bengio},
  \bibinfo{person}{H.~Wallach}, \bibinfo{person}{R.~Fergus},
  \bibinfo{person}{S.~Vishwanathan}, {and} \bibinfo{person}{R.~Garnett}}
  (Eds.), Vol.~\bibinfo{volume}{30}. \bibinfo{publisher}{Curran Associates,
  Inc.}
\newblock


\bibitem[Wen et~al\mbox{.}(2023)]%
        {wen-etal:2023}
\bibfield{author}{\bibinfo{person}{Qingsong Wen}, \bibinfo{person}{Tian Zhou},
  \bibinfo{person}{Chaoli Zhang}, \bibinfo{person}{Weiqi Chen},
  \bibinfo{person}{Ziqing Ma}, \bibinfo{person}{Junchi Yan}, {and}
  \bibinfo{person}{Liang Sun}.} \bibinfo{year}{2023}\natexlab{}.
\newblock \showarticletitle{Transformers in Time Series: A Survey}. In
  \bibinfo{booktitle}{\emph{Proceedings of the Thirty-Second International
  Joint Conference on Artificial Intelligence, Survey Track. Macao, 19--25
  August 2023}}.
\newblock
\urldef\tempurl%
\url{https://doi.org/10.24963/ijcai.2023/759}
\showDOI{\tempurl}


\end{thebibliography}
\end{document}